\documentclass{article}

\usepackage{arxiv}

\usepackage[utf8]{inputenc} 
\usepackage[T1]{fontenc}    
\usepackage{hyperref}       
\usepackage{url}            
\usepackage{booktabs}       
\usepackage{amsfonts}       
\usepackage{nicefrac}       
\usepackage{microtype}      
\usepackage{graphicx}
\usepackage{natbib}
\usepackage{doi}

\usepackage{booktabs}
\usepackage[title]{appendix}
\usepackage[ruled,linesnumbered]{algorithm2e}
\usepackage{graphicx}
\graphicspath{{./images/}}
\usepackage{subcaption}

\usepackage{booktabs}  
\usepackage{multirow}  
\usepackage{makecell}
\usepackage{fancyhdr}

\title{LimSim Series: An Autonomous Driving Simulation Platform for Validation and Enhancement}

\date{} 					

\author{
{\bf Daocheng Fu$^{1\dagger}$, Naiting Zhong$^{2,1\dagger}$, Xu Han$^{3,1\dagger}$, Pinlong Cai$^{1\spadesuit}$, Licheng Wen$^{1}$,  Song Mao$^{1}$, } \\ \\
{\bf Botian Shi$^{1}$, Yu Qiao$^{1}$} \\ \\
$^\dagger$~Equal Contributors \; 
$^{\spadesuit}$~Corresponding Author ( \texttt{caipinlong@pjlab.org.cn})\\ \\
$^{1}$~Shanghai Artificial Intelligence Laboratory, Shanghai, China  \; 
$^{2}$~Tongji University, Shanghai, China  \\ \\ 
$^{3}$~The Hong Kong University of Science and Technology (Guangzhou), Guangzhou, China
}

\begin{document}
\maketitle
\begin{abstract}
	
Closed-loop simulation environments play a crucial role in the validation and enhancement of autonomous driving systems (ADS). However, certain challenges warrant significant attention, including balancing simulation accuracy with duration, reconciling functionality with practicality, and establishing comprehensive evaluation mechanisms. This paper addresses these challenges by introducing the LimSim Series, a comprehensive simulation platform designed to support the rapid deployment and efficient iteration of ADS. The LimSim Series integrates multi-type information from road networks, employs human-like decision-making and planning algorithms for background vehicles, and introduces the concept of the Area of Interest (AoI) to optimize computational resources. The platform offers a variety of baseline algorithms and user-friendly interfaces, facilitating flexible validation of multiple technical pipelines. Additionally, the LimSim Series incorporates multi-dimensional evaluation metrics, delivering thorough insights into system performance, thus enabling researchers to promptly identify issues for further improvements. Experiments demonstrate that the LimSim Series is compatible with modular, end-to-end, and VLM-based knowledge-driven systems. It can assist in the iteration and updating of ADS by evaluating performance across various scenarios. The code of the LimSim Series is released at: \href{https://github.com/PJLab-ADG/LimSim}{https://github.com/PJLab-ADG/LimSim}.
\end{abstract}


\section{Introduction}

Validating and enhancing autonomous driving systems (ADS) within closed-loop simulation environments has become a critical focus of recent intelligent transportation research \citep{gulino2024waymax,ljungbergh2025neuroncap,yang2024drivearena}. Such environments offer a continuous cycle of data collection, model training, and performance evaluation, expanding the capabilities of ADS by providing invaluable feedback loops \citep{codevilla2019exploring}. From a temporal perspective, closed-loop environments enable the exploration of long-term performance of decision-making and planning. Spatially, they offer diverse and dynamic scenarios that help uncover corner cases. In terms of control continuity, closed-loop simulation evaluates the interaction of different system modules, thereby revealing potential weaknesses. This makes closed-loop simulation indispensable in the development and refinement of ADS \citep{zhang2022rethinking}. However, building an effective closed-loop autonomous driving simulation must strike a balance between realism, system requirements, and performance \citep{fu2024limsim++}. Achieving this requires not only designing realistic driving scenarios but also meeting the diverse and evolving needs of ADS technologies. Current autonomous driving simulation platforms still face several key challenges.

First, \textbf{Difficulty in Balancing Simulation Accuracy and Duration.} Although existing simulators 
have made significant strides in realism, providing diverse and complex driving scenarios, these simulators often face the challenge of balancing simulation accuracy with duration. Real-time performance constraints require compromises in accuracy, which can undermine the effectiveness of the simulation \citep{gog2021pylot}. For example, vehicle-based simulators like CARLA \citep{dosovitskiy2017carla} and AirSim \citep{shah2018airsim} typically provide rigid control over background vehicles, limiting the realism of traffic interactions. On the other hand, data-driven simulators such as SimNet \citep{bergamini2021simnet} and TrafficGen \citep{feng2023trafficgen}, which rely on real-world driving data, struggle to scale for large-scale, long-duration simulations. This trade-off between accuracy and simulation efficiency diminishes the reliability of the validation process, making it difficult to simulate real-world scenarios that call for both precision and long-term performance assessment.

Second, \textbf{Conflict between Functionality and Practicality.} As autonomous driving research progresses, ADS architectures undergo multiple iterations, resulting in modular \citep{zhu2021survey}, end-to-end \citep{chen2024end}, and knowledge-driven technical pipelines \citep{li2023towards}. These pipelines often have specific and varied requirements for inputs and outputs during the validation process. When developers aim to test and refine their systems using existing simulators,
they frequently encounter challenges related to these simulators' inflexibility. For example, these platforms may struggle to adapt algorithms and manage upstream and downstream data in ways that align with different pipeline architectures. This often necessitates considerable time and effort to configure the environment, which can be especially cumbersome when testing individual modules like decision-making algorithms that may not independently control a vehicle \citep{fu2024limsim++}. To address this issue, providing baseline algorithms that can serve as placeholders or default solutions for missing components can streamline testing. While platforms like Apollo and Autoware offer comprehensive baseline algorithms and modular designs, they are often too complex for general researchers due to steep learning curves and high access barriers \citep{hallyburton2023avstack,ochs2024one}. Thus, these platforms slow down the iteration process, which delays the advancement of ADS development.

Third, \textbf{Lack of a Comprehensive and Reasonable Evaluation System.} Traditional open-loop evaluation systems assess individual tasks in the autonomous driving pipeline, such as measuring perception accuracy using metrics like Average Precision with Heading (APH) and Intersection over Union (IoU) \citep{feng2020deep}, or evaluating trajectory prediction with Average Displacement Error (ADE) and Final Displacement Error (FDE) \citep{huang2022survey}. While these metrics are valuable for assessing specific modules, they fail to offer a holistic evaluation of the entire ADS. In contrast, closed-loop simulation enables continuous sampling in a data space, which requires a corresponding evaluation system capable of handling continuous data inputs. Most current closed-loop systems evaluate performance using metrics such as human-driving similarity, traffic rule violations, and goal achievement \citep{caesar2021nuplan}. However, these metrics have notable limitations. For instance, they may fail to capture the root causes of issues, such as errors in earlier stages of decision-making that lead to collisions later in the simulation. A robust evaluation system must be able to identify key moments and scenes where performance deviates from expectations, enabling developers to pinpoint problematic modules \citep{li2023survey}. Furthermore, for systems with multiple interacting components, the evaluation system should be capable of assigning responsibility to the correct module to ensure accurate performance metrics \citep{li2024data}. Thus, a comprehensive evaluation framework is essential for the continuous improvement of ADS, ensuring that all modules are properly tested and refined.

\begin{figure*}
    \centering
    \includegraphics[width=1.0\linewidth]{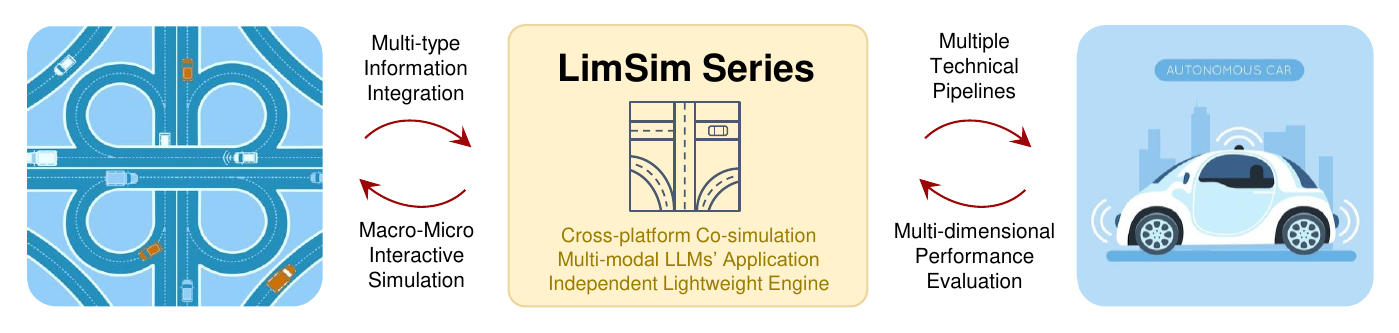}
    \caption{\textrm{The LimSim Series is all you need for agile and effective autonomous driving simulation.}}
    \label{fig:intro}
\end{figure*}

To address these challenges and further the validation and enhancement of ADS in closed-loop environments, we developed the LimSim Series, a comprehensive simulation platform. As illustrated in Figure~\ref{fig:intro}, through the integration of multi-type information from road network, the LimSim Series controls background vehicles through human-like decision-making and planning algorithms, creating realistic driving scenarios for algorithm testing. This approach balances simulation efficiency with quality by introducing the concept of the Area of Interest (AoI), which optimizes computational resources and enables macro-micro interactive simulation. The LimSim Series also offers a variety of baseline algorithms and user-friendly interfaces, facilitating the rapid deployment of diverse ADS and algorithms. These features support flexible validation of multiple technical pipelines, allowing researchers to efficiently test different modules and configurations. Moreover, the LimSim Series incorporates multi-dimensional evaluation metrics, which provide detailed insights into system performance, helping researchers quickly identify issues and areas for improvement. 

The contributions of this paper are summarized as follows:

\begin{itemize}
    \item Integrated framework for ADS pipelines and closed-Loop simulation: We delve into an in-depth analysis of the key challenges in the development of closed-loop simulation environments, highlighting the importance of an ADS simulation platform that supports multi-source inputs and diverse technological pipelines, thereby proposing a comprehensive simulation framework.

    \item Development of the LimSim Series: We introduce the open-source autonomous driving simulation platform designed to support rapid deployment and efficient iteration. With its modular architecture and built-in baseline algorithms, the LimSim Series enables flexible validation of multiple ADS pipelines. The platform’s comprehensive evaluation system further aids in the continuous improvement of autonomous driving solutions.

    \item Extensive experimental validation: We conduct extensive experiments to test different technical pipelines with multi-modal inputs, demonstrating how the LimSim Series can effectively validate and enhance ADS performance and highlighting its potential to accelerate the development of reliable and efficient ADS.
\end{itemize}

\section{Related Work}

\label{section:ads}

In recent years, autonomous driving technology has grown rapidly. With quick advancements and new solutions, various ADS have emerged and continue to evolve \citep{zhao2024autonomous}. As these systems become more complex, the methods for testing them are also changing \citep{li2024choose}. To create a simulation platform that meets developing needs, we have studied the main ADS technical pipelines, especially from the validation perspective, and explored effective ways to test and improve their performance.

\begin{figure*}
    \centering
    \includegraphics[width=0.85 \linewidth]{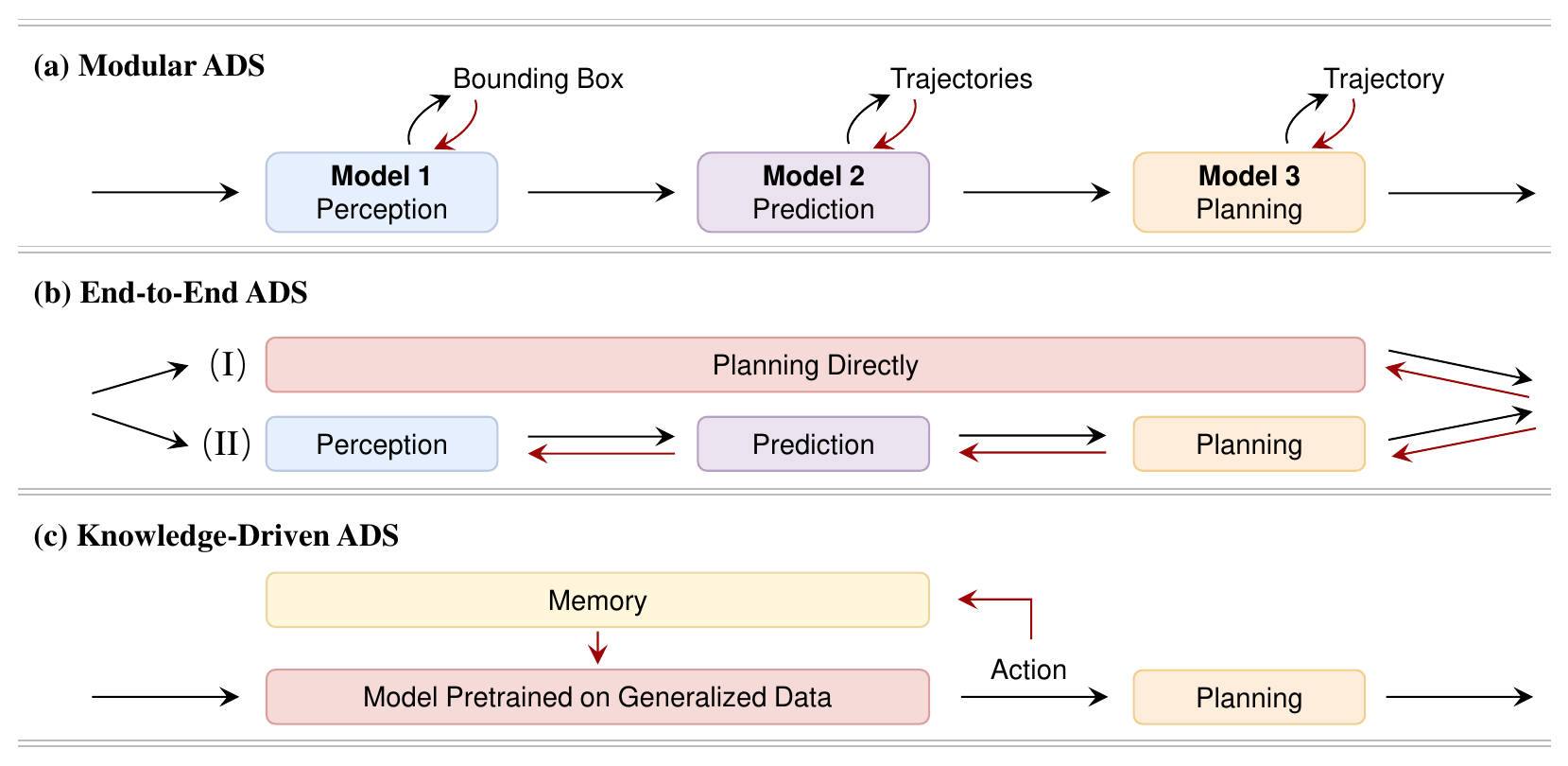}
    \caption{\textrm{Multiple technology pipelines for autonomous driving systems.}}
    \label{fig:ads_system}
\end{figure*}

\subsection{ADS Pipelines}

As shown in Figure~\ref{fig:ads_system}, mainstream ADS  can be categorized into three types of technical pipelines: Modular ADS, End-to-end ADS, and Knowledge-driven ADS \citep{chen2024end}. The following is a detailed introduction.

\subsubsection{Modular ADS}

The development of ADS has been significantly influenced by machine learning \citep{huang2020autonomous}. In the early stages, limitations in computing power, neural network size, and the scale of training data prevented a single model from handling the entirety of autonomous driving tasks. To address these constraints, researchers divided the overarching task into multiple sequential sub-tasks, enabling focused development on specific aspects of the system \citep{chen2015deepdriving,maddern20171,bachute2021autonomous}. This modular approach facilitated rapid industrial adoption of ADS technologies, which in turn attracted substantial investment and accelerated progress in the field.

As depicted in Figure~\ref{fig:ads_system}, modular ADS consists of multiple interconnected modules, typically categorized into perception, prediction, and planning \citep{mozaffari2020deep,kiran2021deep}. Some designs integrate prediction and planning into a single module and add a separate control module \citep{zhu2021survey}. Each module operates independently and is trained with its own task-specific loss function \citep{grigorescu2020survey}. This design simplifies practical implementation by isolating tasks, but it also introduces challenges. Fragmentation makes error tracing difficult and can lead to error accumulation across modules. Furthermore, requiring different models for each task reduces computational efficiency and increases the risk of local optima \citep{chib2023recent}.

Another key challenge is evaluation. Testing platforms must provide appropriate inputs for each sub-model and establish an evaluation system capable of attributing responsibility for errors accurately \citep{chao2020survey}. Despite these challenges, modular ADS remains widely used in industry due to its robustness and adaptability. Many modular algorithms and models are expected to persist as redundant components in future systems, serving as safety-enhancing fallbacks \citep{yurtsever2020survey}.

\subsubsection{End-to-End ADS}

Attempts at end-to-end ADS began as early as 1988 with ALVINN, a system that employed a shallow neural network to drive a vehicle \citep{pomerleau1988alvinn}. However, early systems faced severe limitations in generalization and were ineffective in complex traffic scenarios. Over time, advances in computational power and algorithmic sophistication enabled significant progress in end-to-end approaches. Modern end-to-end ADS leverages a single integrated model to perform all driving tasks \citep{zeng2019end,tampuu2020survey,chitta2021neat}. While it may still employ a modular structure, the entire model is trained jointly, optimizing for the ultimate driving task \citep{le2022survey}. This integrated approach offers several advantages. By reducing system complexity and improving computational and training efficiency, end-to-end ADS can adapt more readily to complex traffic conditions\citep{hu2023planning}. Joint training also mitigates issues of error propagation between modules, a common problem in modular systems \citep{casas2021mp3,chib2023recent}.

Nevertheless, end-to-end ADS faces unique challenges, particularly in testing. Open-loop testing often fails to capture real-world complexities, as predictions on test datasets closely match ground truth, resulting in high scores \citep{caesar2021nuplan,jia2024bench2drive}. In contrast, real-world driving involves continuous decision-making, where small errors accumulate over time and unanticipated events challenge the system's robustness \citep{zhang2022rethinking,chen2024end}. Consequently, realistic, interactive, and editable closed-loop testing environments are crucial for accurately evaluating the performance of end-to-end ADS.

\subsubsection{Knowledge-Driven ADS}

Despite significant advancements in autonomous driving, the diversity and complexity of road environments continue to create endless corner cases \citep{bolte2019towards}. A common strategy to address this issue is collecting more data for training \citep{li2022coda}. However, the sheer variability of driving scenarios renders data collection an unending task. The root of the problem lies in the separation of model training and deployment. Unlike human drivers, whose skills improve with experience, traditional ADS models lack mechanisms for continuous learning \citep{lan2022instance,wen2023dilu}.

Knowledge-driven ADS seeks to bridge this gap by adopting a design that enables continuous learning during operation \citep{mao2023gpt,li2023towards,xu2024drivegpt4,tang2024grounded}. This approach draws inspiration from human drivers, who not only learn vehicle operation and traffic rules but also refine their skills through real-world experience. Leveraging pre-trained models with advanced knowledge and reasoning capabilities, such as large language models (LLMs), knowledge-driven ADS systems can explore environments, respond to unexpected situations, and accumulate experience through reflective learning \citep{sha2023languagempc,cui2023drivellm,wen2023dilu,mei2024continuously}.

So far, the effectiveness of knowledge-driven ADS has primarily been demonstrated in simulation environments \citep{jin2023surrealdriver,chen2024driving,cui2024drive}. This design paradigm, however, introduces promising insights for future ADS development. Like end-to-end systems, knowledge-driven ADS requires high-quality closed-loop testing environments to support interaction and the accumulation of driving experience. Additionally, pre-trained models often rely on generic data, which can lead to format mismatches with autonomous driving-specific datasets \citep{cui2024survey}. Testing platforms must therefore provide robust tools for format conversion, control interfaces, and auxiliary functionalities to fully realize the potential of knowledge-driven ADS.

The distinct advantages and unique challenges of modular, end-to-end, and knowledge-driven ADS methodologies highlight the critical need for robust simulation and testing platforms, which remain pivotal for advancing ADS toward widespread adoption.

\subsection{Validation and Enhancement of ADS}

To accurately assess performance at various stages of ADS development, testing methods are adjusted accordingly. In this section, we survey and compare widely used testing methods and summarize the requirements for building a closed-loop autonomous driving testing platform.

\subsubsection{Dataset and Benchmark}

Creating datasets and benchmarks is a fundamental approach for training and evaluating autonomous driving models. These datasets are typically constructed by collecting data from vehicle-mounted or roadside sensors, which are then labeled according to the specific requirements of the task. Well-known autonomous driving datasets include KITTI \citep{geiger2013vision}, BDDV \citep{xu2017end}, HDD \citep{ramanishka2018toward}, nuScenes \citep{caesar2020nuscenes}, and Waymo Open Dataset \citep{sun2020scalability}. These datasets are gathered using a variety of sensors, such as cameras, LiDAR, radar, and GPS, which capture data on traffic participants and their behaviors. The labeled data serves as the foundation for training models, enabling the advancement of cutting-edge algorithms in the field of autonomous driving \citep{li2024ego,wang2024driving,zheng2025genad}. Due to the significant cost and labor involved in the collection of real-world data, many research teams rely on these widely used, open-source datasets, which help establish baseline standards for model performance \citep{feng2020deep,liu2024survey}. Additionally, these datasets often come with predefined evaluation metrics and benchmarks, which are critical for comparing the performance of different models across a variety of tasks \citep{guo2019safe}.

Open-loop datasets, which are directly sourced from real-world driving environments, provide authentic training samples and create a shared platform for models to compete on. This has been a significant advancement in autonomous driving research, enabling researchers to test their models against the same set of real-world data. However, while these datasets offer valuable insights, they have limitations. Specifically, open-loop datasets do not provide feedback on model outputs; they simply measure the discrepancy between model predictions and the ground-truth labels \citep{codevilla2019exploring,caesar2021nuplan}. This creates a gap in fully assessing the model’s actual performance in real-world conditions. The issue is twofold: first, a small gap between model output and labels indicates that the model is feasible but does not necessarily imply optimal performance. Second, task-specific evaluation metrics do not always correlate with the broader success of an ADS \citep{zhang2022rethinking,ljungbergh2025neuroncap,yang2024drivearena}. A model may perform well on a specific task but still fail to operate effectively in the context of a fully integrated ADS.

Moreover, the high cost of real-world data collection limits the scalability of datasets, preventing researchers from capturing rare or complex driving scenarios \citep{chitta2022transfuser,shao2023safety}. This limitation slows the development of more robust ADS technologies. In conclusion, while open-loop datasets and benchmarks are critical for advancing research, their static nature and the challenges of real-world data collection highlight the need for new approaches, such as simulation-based data generation, to overcome these barriers and improve the generalization of autonomous systems \citep{hu2023simulation,tian2024robust}.

\subsubsection{Simulation}

Autonomous driving simulators can be roughly categorized into flow-based, vehicle-based, and data-based \citep{wenl2023limsim}.

Flow-based simulation systems have been developed over the years to assist urban planners and traffic managers. Notable examples include PARAMICS \citep{cameron1996paramics}, a commercial software released in 1998, which integrates traffic simulation, visualization, road network design, and adaptive signal control. Vissim \citep{fellendorf2010microscopic}, another commercial tool, offers high-level visualizations of traffic scenarios using realistic models. CORSIM \citep{halati1997corsim}, supported by the Federal Highway Administration, specializes in road geometry, traffic control, and large-scale simulations. Aimsun \citep{barcelo2005dynamic} is a widely used software for traffic planning and demand analysis, while SUMO is an open-source tool for modeling urban traffic and intermodal transportation systems, offering features like route planning and emission calculations. Despite their strength in simulating large-scale traffic networks, flow-based simulators use simple car-following models, limiting their ability to accurately capture detailed vehicle behavior and microscopic movements \citep{kotusevski2009review}.

Vehicle-based simulators provide more dynamic and realistic simulations by focusing on vehicle-specific behaviors \citep{gog2021pylot,xu2023opencda,bockman2024aark}. Early autonomous driving simulators, such as USARSim \citep{carpin2007usarsim} and Webots \citep{michel2004cyberbotics}, utilized modified game engines to simulate physical interactions. Today, simulators like Gazebo \citep{koenig2004design}, AirSim, LGSVL \citep{rong2020lgsvl}, and CARLA offer advanced features. Gazebo is a flexible 3D simulator often used with ROS \citep{quigley2009ros} for dynamic rendering and object interaction. AirSim, based on Unreal Engine \citep{sanders2016introduction}, provides high-fidelity vehicle simulations and a variety of urban scenarios. LGSVL, built on Unity \citep{haas2014history}, supports detailed sensor simulations like LiDAR and radar. CARLA, also open-source, offers customizable environments and sensor suites for autonomous driving research. Vehicle-based simulators allow for precise vehicle motion modeling and decision-making algorithm testing. However, they fall short in simulating realistic background traffic flow, as they lack the capacity for dynamic vehicle interaction on a larger scale \citep{wenl2023limsim}.

Data-based simulators leverage traffic flow data to simulate realistic driving scenarios \citep{hallyburton2023avstack}. These systems generate vehicle behavior based on historical data, making it challenging to model interactions between the ego vehicle and other road users. SimNet, the first machine learning-based simulator, generates realistic driving episodes from historical data and improves as more data is used. InterSim \citep{sun2022intersim} and TrafficGen \citep{feng2023trafficgen}, both data-driven systems, simulate vehicle interactions and generate diverse traffic scenarios. Data-based simulators are effective at learning multi-vehicle interactions from real-world data, but they are highly dependent on the available dataset, which limits their ability to create new, diverse scenarios and can lead to fragmented simulations \citep{mutsch2023model}.

In general, a simulation platform that can efficiently generate detailed scenarios while balancing real-time performance and simulation fidelity, and readily support the creation of long-term dynamic traffic data and convenient in-depth evaluation, is crucial for advancing autonomous driving development.

\section{Comprehensive Simulation Framework}
\begin{figure*}
    \centering
    \includegraphics[width=0.85\linewidth]{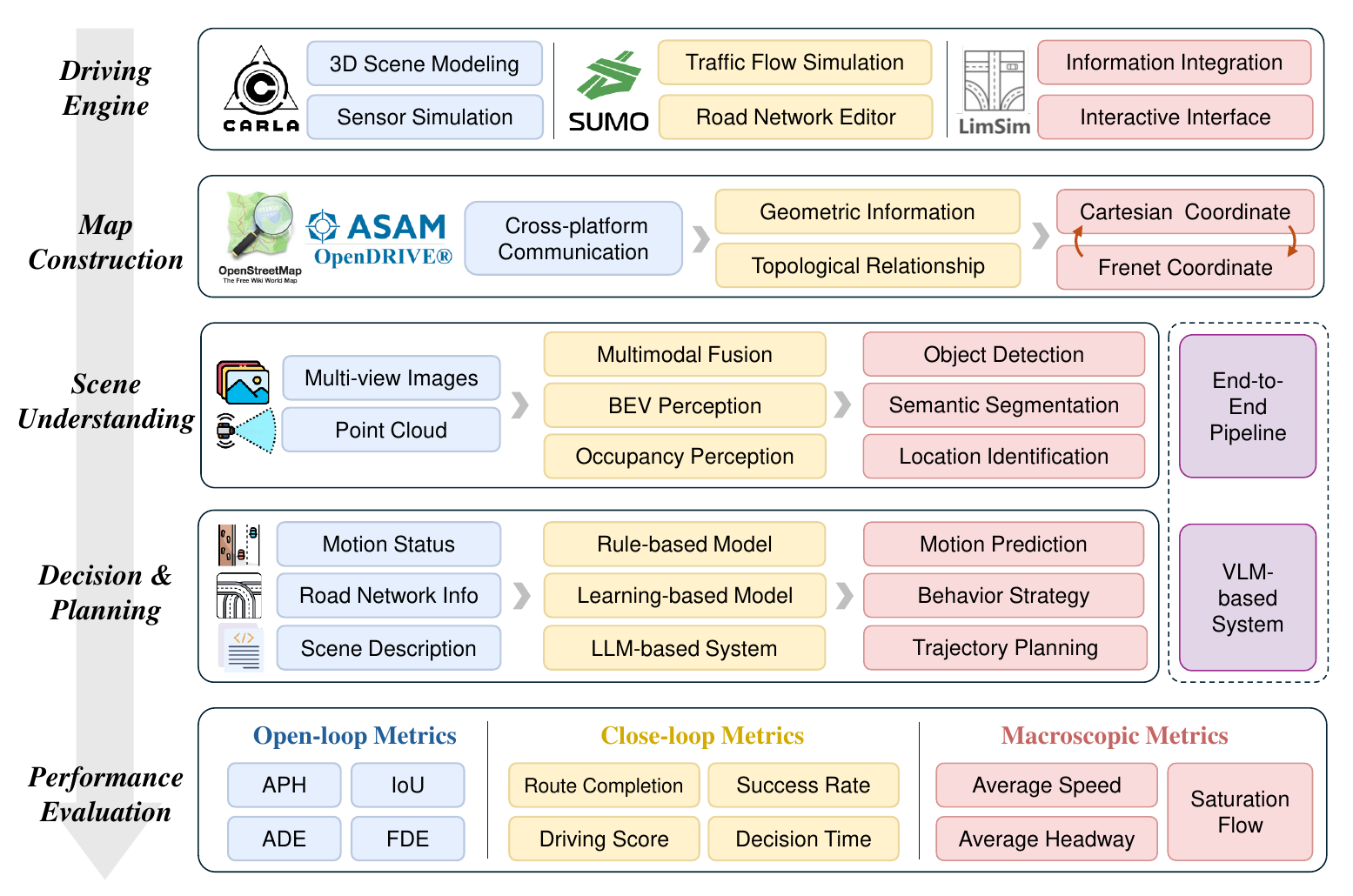}
    \caption{
    \textrm{(1) \textbf{Driving Engine} emphasizes independence and compatibility, allowing for full deployment and testing of algorithms while interfacing with open-source engines like SUMO and CARLA for cross-platform development. (2) \textbf{Map construction} involves importing OpenDrive-formatted files or obtaining data through cross-platform communication, utilizing Cartesian and Frenet coordinate systems for accurate vehicle positioning and trajectory planning. (3)\textbf{ Scene understanding }is enhanced through 3D scene information from CARLA, enabling the use of sensor data for traffic participant identification and motion state estimation, which can be applied to perception algorithms or integrated into end-to-end architectures. (4)\textbf{Decision and planning} include traditional models and advanced joint decision planning models that leverage MCTS for behavior decisions and parallel trajectory planning for trajectory generation. (5) \textbf{Performance evaluation} is facilitated through a series of indicators focusing on vehicle operation status, with the ability to calibrate model parameters to align with real-world data for enhanced simulation realism.}
}
    \label{fig:framework}
\end{figure*}

This paper provides a comprehensive framework for autonomous driving simulation, which includes multiple necessary modules as shown in Figure~\ref{fig:framework}, to support the deployment of different technology pipelines. We then share some considerations and insights from designing this framework.

\subsection{Simulation Modules}
\subsubsection{Driving Engine}
In the LimSim Series, we have paid great attention to the issues of independence and compatibility with existing simulation engines. An independent simulation engine is crucial because it allows users to fully deploy and test algorithms based on this platform, effectively obtaining all dynamic and static information of the simulation process. This complete access is essential for evaluating ADS in a controlled setting. Moreover, the LimSim Series is designed to seamlessly integrate with widely used open-source simulation engines such as SUMO and CARLA, enhancing its versatility. By offering cross-platform communication capabilities, the LimSim Series allows for joint development, enabling users to leverage the strengths of each platform. For instance, SUMO is known for its high-speed traffic flow simulation, which is ideal for large-scale traffic management studies, while CARLA excels in realistic 3D rendering, making it perfect for visualizing complex urban environments. Through the integration of these platforms, the LimSim Series can support a wide variety of autonomous driving technology pipelines, enabling users to test and validate their algorithms under diverse simulation scenarios.

\subsubsection{Map Construction}

Traffic network map construction in the LimSim Series is designed to be flexible and user-friendly, with support for importing map files in formats such as the  ASAM OpenDRIVE format. Additionally, users can obtain map information through cross-platform communication, which ensures the adaptability of the platform to various data sources. The geometric information and topological relationships of the network are fundamental in the simulation process, helping to establish effective relationships between vehicles and roads, thereby indirectly constructing relationships between vehicles and their surrounding traffic participants. Moreover, to accurately locate vehicle positions, the LimSim Series incorporates a dual-coordinate system, consisting of the Cartesian and Frenet coordinate systems. The Cartesian coordinate system provides the absolute positioning of vehicles, which is essential for analyzing vehicle conflicts and assessing driving performance. For example, by using Cartesian coordinates, users can track how a vehicle responds when changing lanes or approaching an intersection. On the other hand, the Frenet coordinate system is particularly useful for local trajectory planning. It simplifies the generation of trajectories close to the road centerline, without the need for detailed road geometry. This dual approach ensures precise localization and effective path planning, even in complex environments.

\subsubsection{Scene Understanding}
Traffic scenes can be defined as the road areas surrounding the target vehicle or target road points, including all traffic participants within a specified range. By exchanging information with 3D scene simulators such as CARLA, the LimSim Series provides detailed 3D road scene data, which includes raw sensor data from multiple perspectives—such as images and point clouds from cameras and LiDAR sensors. These data can be used with traditional perception algorithms to identify and tag traffic participants in the scene, as well as estimate their motion states (e.g., speed and direction). Furthermore, this sensor data can also be fed into end-to-end deep learning architectures for implicit encoding of the scene, or into multimodal large models for general context understanding, enabling more sophisticated scene interpretation.

\subsubsection{Decision and Planning}
The LimSim Series offers a range of trajectory planning methods, catering to both simple and advanced use cases. The baseline methods, including traditional following and lane-changing models, serve as a starting point for users new to autonomous driving simulation. For instance, the combination of the Intelligent Driver Model \citep{treiber2000congested} and the MOBIL model \citep{kesting2007general} provides a simple yet effective approach for simulating vehicle movement in traffic. These models help users understand the fundamental mechanics of vehicle behavior, such as maintaining safe following distances and executing smooth lane changes. Although the control mode is straightforward, these models also serve as a baseline for evaluating traffic flow simulation performance. For more complex scenarios, LimSim includes a joint decision planning model that employs a two-layer logic structure. The upper layer utilizes Monte Carlo Tree Search (MCTS) \citep{browne2012survey} for decision-making, which helps to model the vehicle's high-level behavior, such as determining whether to overtake another vehicle or stop at an intersection. The lower layer performs trajectory planning, generating specific trajectory points using parallel processing to ensure real-time responsiveness. Additionally, advanced techniques such as LLMs can be integrated to output decision meta-actions based on scene analysis. For example, an LLM could interpret the surrounding traffic conditions and decide whether the vehicle should accelerate, decelerate, or change lanes.

\subsubsection{Performance Evaluation}
To evaluate the performance of algorithms, the LimSim Series introduces a comprehensive set of performance indicators. For individual vehicles, closed-loop simulation indicators mainly focus on the vehicle's operating status, including efficiency, comfort, and safety indicators. For example, efficiency could be measured by the completion time of the driving task, comfort could be assessed through the smoothness of the vehicle’s ride, and safety could be evaluated by the frequency of collisions or near-misses. The quantification of these indicators can help users identify algorithmic failures in certain scenarios, thereby identifying corner cases and supporting algorithm iteration and optimization. For instance, a scenario where a vehicle repeatedly fails to make safe lane changes might highlight the need for improving the vehicle’s decision-making or perception models. In addition, to further enhance the realism of the simulation, users can calibrate model parameters to match the statistical characteristics of real-world datasets, such as average speed, headway (the time gap between two vehicles), and saturation flow (the maximum number of vehicles that can pass through an intersection in a given time). This ensures that the simulation results are representative of real-world traffic conditions, enhancing the validity of the testing environment.

\subsection{Design Considerations and Insights}

\subsubsection{Trade-off Between Efficiency and Scale}

\begin{figure*}
    \centering
    \includegraphics[width=1.0\linewidth]{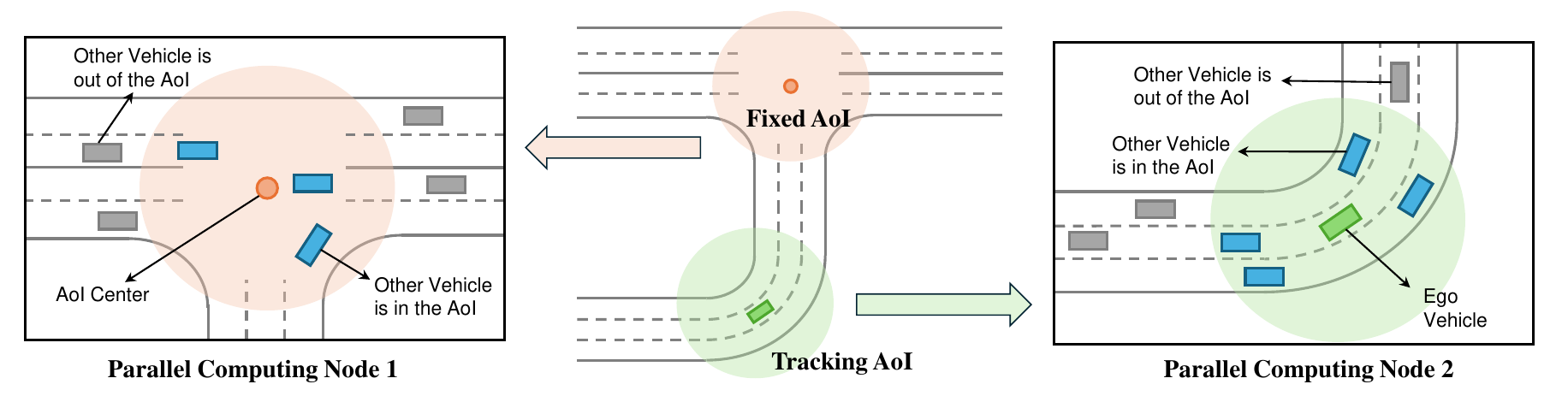}
    \caption{\textrm{Differentiated decision-making planning strategies for other vehicles inside and outside the AoI.}}
    \label{fig:AoI}
\end{figure*}

As the number of vehicles operating on the platform increases, the computational load required to simulate their interactions also grows exponentially. With limited computing resources, a common challenge faced by simulation platforms is the trade-off between efficiency and scale. In simpler terms, the more vehicles and detailed interactions you want to simulate, the more computational power is needed. In existing simulators, this trade-off is often addressed by reducing simulation granularity—this can include lowering the rendering frame rate or decreasing the frequency of trajectory updates. However, such simplifications may sacrifice realism and accuracy, especially in critical scenarios where vehicle behavior is key.

To balance these conflicting demands, we introduce the concept of the Area of Interest (AoI) in the LimSim Series, as illustrated in Figure~\ref{fig:AoI}. The AoI focuses computational resources on a localized region around the vehicle of interest, ensuring that the simulation remains highly detailed where necessary, and more computationally efficient in less critical areas. Inside the AoI, vehicle behavior is simulated with high granularity, using complex control strategies that mimic real-world decision-making. For example, when a vehicle is about to merge into another lane, it will apply a nuanced control strategy to assess and react to nearby traffic. Outside the AoI, simulation granularity is reduced, and vehicles default to simple, high-efficiency behaviors such as lane-following or basic speed control. This approach allows for a scalable solution where the simulator can handle a large number of vehicles without overwhelming the system. In the future, we also plan to explore distributed simulation strategies, where the control of vehicles could be decentralized and managed by a network of distributed centers. This would allow for greater flexibility and scalability, while maintaining a unified scene rendering interface to ensure consistent visual representation across all simulation modules.

\subsubsection{Integration of Real Data with Simulation Platform}

One of the critical challenges in simulation-based research is integrating real-world data to improve the realism of the simulated environment. Importing real traffic data, such as road network layouts and historical traffic flow, enhances the accuracy of the simulated traffic conditions. However, this integration can lead to the loss of dynamic interaction between vehicles, which is essential for simulating realistic behavior.

\begin{figure*}
    \centering
    \includegraphics[width=1.0\linewidth]{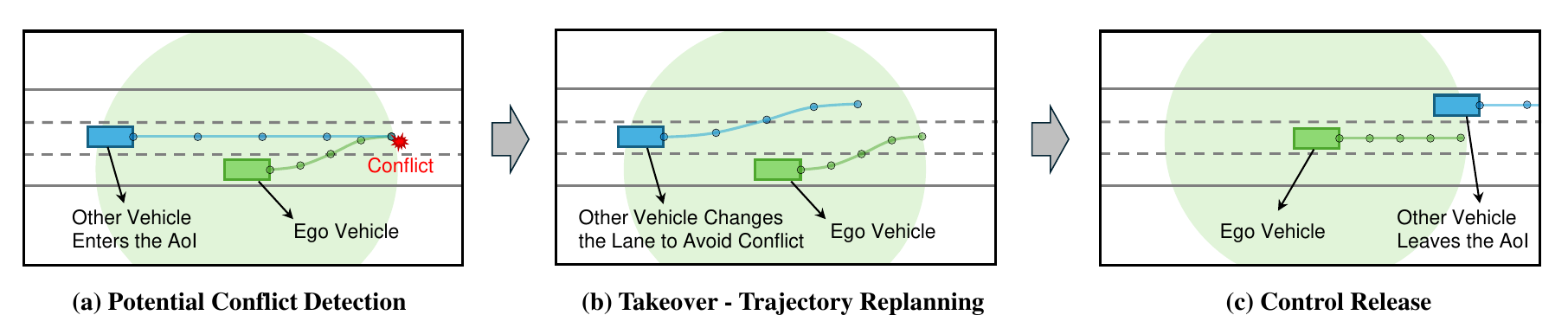}
    \caption{\textrm{Interactive simulation strategy combines virtual simulation and real traffic data.}}
    \label{fig:control}
\end{figure*}

To address this issue, we have developed interactive simulation strategies as a core feature of the LimSim Series. These strategies help determine when vehicles should be controlled by the simulator and when they should follow real-world data. This dynamic approach enhances both realism and interactivity, allowing the simulation to adapt to the complexities of traffic dynamics. For example, as shown in Figure~\ref{fig:control}, when a vehicle enters the AoI and may potentially conflict with others (such as causing a rear-end collision), the simulator will override the vehicle's trajectory with a default control strategy to prevent accidents or unnatural behavior. Once the conflict is resolved or the vehicle exits the AoI, the simulator will restore the vehicle's trajectory to align with real-world data. This ensures that while the vehicle's actions are dynamically adjusted to avoid collisions, they remain grounded in the real-world traffic patterns outside the AoI. In our previous work, we also explored the integration of 3D scene reconstruction with the LimSim Series \citep{yan2024oasim}. This combination allows us to replicate real-world traffic environments in greater detail and provides the flexibility to edit vehicle trajectories arbitrarily. Such an integration opens up new possibilities for simulating complex urban driving scenarios.

\section{Experiments}
\label{section:experiments}

\subsection{Experiment setting}

The LimSim Series provides a variety of baseline modules and algorithms, along with user-friendly APIs, enabling seamless integration with mainstream autonomous driving systems. It offers convenient simulation and validation functionalities for different autonomous driving systems, helping to explore their performance boundaries. Additionally, the LimSim Series fully supports high-definition map parsing, allowing for simulation experiments across various road types and scenarios.

In this section, we conducted simulation experiments on different types of autonomous driving systems in diverse scenarios. The systems evaluated include: (1) \textit{PDM} \citep{Dauner2023CORL}, representing modular autonomous driving systems; (2) \textit{Interfuser} \citep{shao2022interfuser}, representing end-to-end autonomous driving systems; (3) \textit{VLM-Agent} \citep{wen2023dilu, fu2024limsim++}, representing knowledge-driven autonomous driving systems; and (4) \textit{LimSim-TM} \citep{wenl2023limsim}, the baseline traffic controller provided by the LimSim Series. PDM includes several modules such as agent forecast, trajectory proposal, and trajectory refinement to provide appropriate trajectories for vehicles. It uses a rule-based predictive planner to obtain a trajectory proposal, and a learned ego-forecasting module to refine the trajectory. InterFuser is a security-enhanced autonomous driving strategy based on multi-sensors and integrated with the transformer-based method, using interpretable features to increase the safety of autonomous driving. VLM-Agent utilizes the GPT-4o for autonomous driving decision-making with the zero-shot approach. The model performs scenario analysis, behavior prediction, and action decision based on surround-view images provided by the LimSim Series. The decision results of the model will be parsed by the LimSim Series and ultimately applied to the ego car. LimSim-TM uses several different modules to achieve the functions of prediction, decision-making, and planning through search. The decision-making module introduces social value orientation (SVO) grouped decision-making, making vehicle behavior closer to real-world situations.

\begin{figure*}
    \centering
    \includegraphics[width=1.0\linewidth]{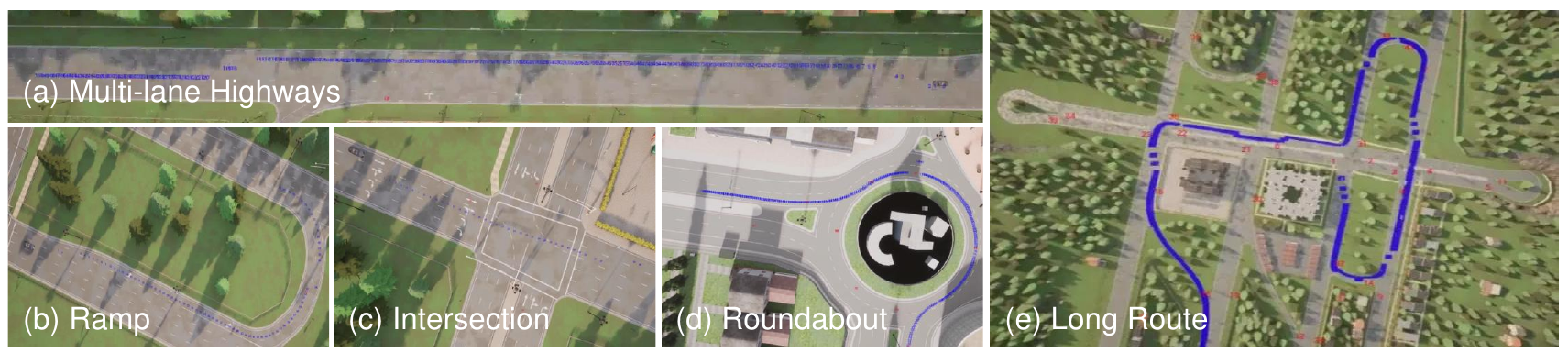}
    \caption{\textrm{Several representative scenarios for performance evaluation of various autonomous driving models.}}
    \label{fig:test_scenario}
\end{figure*}

We selected several representative scenarios for these experiments: a multi-lane highway, a ramp, an intersection, a roundabout, and a custom-designed long route that integrates multiple complex situations. The bird's-eye views of these scenarios, as depicted in Figure \ref{fig:test_scenario}, were captured using CARLA.

\subsection{Performance Evaluation of Various Autonomous Driving Models}

To comprehensively evaluate the performance of autonomous driving systems in the aforementioned scenarios, we assess the simulation results from four metrics: route completion (\%), driving score, average decision time (s), and success rate. The driving score is a holistic measure that takes into account ride comfort, driving efficiency, and safety. For detailed definitions and parameter values, please refer to our previous work \citep{fu2024limsim++}. In each scenario, we generate 10 random background traffic flows to validate the models' performance under various traffic conditions. The success rate is calculated as the number of successful tests out of these 10 experiments. The mean and standard deviation of the experimental results are presented in Table \ref{tab:comparison}.

As the driving scores reflect the comprehensive performance evaluation, LimSim-TM demonstrated the most consistent and superior performance overall. PDM and VLM-Agent showed competitive performance across different scenarios, whereas Interfuser consistently underperformed, especially in roundabout scenarios. The model often drives the vehicle into the middle of the roundabout, causing unsatisfying performance in the roundabout scenario. In contrast, both modular methods and knowledge-driven approaches demonstrated correct decision-making capabilities. This does not imply that end-to-end autonomous driving systems are inherently inferior but rather highlights the sensitivity of data-driven methods to the training data distribution. Since Interfuser was not trained in the selected environment, it could not fully demonstrate its potential. 
In contrast, PDM and LimSim-TM, which rely on rule-based decision-making, can produce relatively strong decisions even in unfamiliar scenarios by employing search-based methods. 

Notably, the zero-shot VLM-Agent has strong generalization abilities even without prior exposure to these scenarios, leveraging its strong commonsense reasoning abilities. However, VLM-Agent suffers from limitations in reasoning speed due to the inference latency of VLM. Its average decision time across various scenarios was approximately 10 seconds, which makes it challenging to meet real-time requirements. Future research could focus on improving the model's inference speed or reducing the number of tokens in the output to bridge the gap between data-driven methods and practical applications.

The experiments demonstrate that the LimSim Series provides a rich simulation environment for different types of autonomous driving systems, enabling diverse interactions and multi-dimensional evaluation of simulation outcomes. The experimental results indicate that the LimSim Series offers evaluations well-suited to the characteristics of various models. Its real-time recoding system also facilitates the identification of corner cases, aiding in model iteration and improvement.

\begin{table}[htbp]
\centering
\rmfamily 
\caption{\textrm{Performance comparison of different models across various scenarios.}}
\begin{tabular}{p{2cm}p{2cm}p{1.8cm}p{1.8cm}p{1.8cm}p{1.8cm}p{1.8cm}}
\toprule
\multirow{2}{*}{Metric} & \multirow{2}{*}{ADS} & \multicolumn{5}{c}{Scenario} \\ \cmidrule{3-7}
& & Highway & Ramp & Intersection  & Roundabout & Long Route \\ 
\midrule
\multirow{4}{*}{\textbf{\makecell{Route \\ Completion \\(\%) $\uparrow$}}} 
& PDM        & 100.00$\pm$0.00 & 94.13$\pm$17.61 & 100.00$\pm$0.00 & 100.00$\pm$0.00 & 84.62$\pm$24.58 \\
& InterFuser & 100.00$\pm$0.00 & 93.18$\pm$20.44 & 100.00$\pm$0.00 & 13.17$\pm$4.90 & 88.14$\pm$25.86 \\
& VLM-Agent & 97.22$\pm$8.33 & 97.82$\pm$6.55 & 100.00$\pm$0.00  & 100.00$\pm$0.00 & 87.90$\pm$19.81 \\
& LimSim-TM & 97.78$\pm$6.67 & 96.98$\pm$9.06 & 100.00$\pm$0.00 & 100.00$\pm$0.00 & 86.15$\pm$28.45 \\
\midrule
\multirow{4}{*}{\textbf{\makecell{Driving \\ Score $\uparrow$}}} 
& PDM        & 40.73$\pm$1.98 & 68.25$\pm$14.01 & 76.30$\pm$2.68 & 27.40$\pm$0.41 & 65.22$\pm$15.04 \\
& InterFuser & 50.99$\pm$4.30 & 54.20$\pm$4.45 & 65.23$\pm$3.85  & 28.00$\pm$16.82 & 37.63$\pm$7.95 \\
& VLM-Agent & 52.98$\pm$9.84 & 47.31$\pm$6.59 & 75.67$\pm$2.14  & 65.40$\pm$9.36 & 30.54$\pm$7.89 \\
& LimSim-TM & 76.28$\pm$18.91 & 64.62$\pm$16.30 & 86.28$\pm$0.51 & 72.08$\pm$14.08 & 78.76$\pm$15.43 \\
\midrule
\multirow{4}{*}{\textbf{\makecell{Avg. \\ Decision \\ Time (s) $\downarrow$}}} 
& PDM        & 0.02$\pm$0.00 & 0.01$\pm$0.00 & 0.01$\pm$0.00 & 0.02$\pm$0.00 & 0.02$\pm$0.00 \\
& InterFuser & 0.11$\pm$0.01 & 0.12$\pm$0.02 & 0.11$\pm$0.00  & 0.12$\pm$0.00 & 0.11$\pm$0.01 \\
& VLM-Agent & 9.68$\pm$1.38  & 10.53$\pm$3.25 & 11.24$\pm$3.36 &  11.70$\pm$3.36 & 11.42$\pm$2.16 \\
& LimSim-TM & 0.09$\pm$0.03  & 0.13$\pm$0.04 & 0.03$\pm$0.00 & 0.05$\pm$0.01 & 0.08$\pm$0.01 \\
\midrule
\multirow{4}{*}{\textbf{\makecell{Success \\Rate $\uparrow$}}} 
& PDM        & 1.00$\pm$0.00 & 0.90$\pm$0.30 & 1.00$\pm$0.00  & 1.00$\pm$0.00 & 0.70$\pm$0.46 \\
& InterFuser & 1.00$\pm$0.00 & 0.90$\pm$0.30 & 1.00$\pm$0.00  & 0.00$\pm$0.00 & 0.80$\pm$0.40 \\
& VLM-Agent & 0.90$\pm$0.30 & 0.90$\pm$0.30 & 1.00$\pm$0.00 & 1.00$\pm$0.00 & 0.70$\pm$0.46 \\
& LimSim-TM & 0.90$\pm$0.30 & 0.90$\pm$0.30 & 1.00$\pm$0.00 & 1.00$\pm$0.00 & 0.80$\pm$0.40 \\
\bottomrule
\end{tabular}
\label{tab:comparison}
\end{table}

\section{Conclusion and Future Work}
\label{section:discussion}

This paper explores the development and validation of ADS, categorizing them into modular, end-to-end, and knowledge-driven approaches. It then introduces the LimSim Series, a comprehensive simulation framework supporting various ADS types through modules like driving engine, map construction, scene understanding, decision and planning, and performance evaluation. Experiments across diverse scenarios demonstrate the ability of the proposed platform to evaluate ADS performance effectively. 

In the future, simulation systems for ADS will need to achieve breakthroughs in the following key areas to meet practical demands.
\begin{itemize} 
\item Support for high-fidelity sensor simulation: Future simulation frameworks could incorporate 3D Gaussian sputtering and diffusion techniques to achieve accurate 3D scene reconstruction, providing more realistic and diverse sensor signal inputs for model testing. Emphasis should be placed on ensuring high rendering efficiency to optimize simulation performance.

\item Simulation of heterogeneous traffic flows: The scenarios involving mixed human-vehicle traffic are common in the road networks, where the interactions between pedestrians and vehicles are simulated. Additionally, the system should accommodate special-purpose vehicles, such as buses, taxis, and ambulances, capturing their unique travel trajectories and behaviors.

\item Comprehensive testing scenario library: A robust simulation platform should offer a diverse and comprehensive set of test scenarios. In addition to scenarios based on log case editing, future research should focus on AI-driven generation methods guided by specific instructions. This approach would be particularly valuable for generating rare and hard-to-collect corner cases, facilitating more thorough testing in complex and edge conditions.
\end{itemize}


\bibliographystyle{cas-model2-names}
\bibliography{references}  

\begin{thebibliography}{92}
\expandafter\ifx\csname natexlab\endcsname\relax\def\natexlab#1{#1}\fi
\providecommand{\url}[1]{\texttt{#1}}
\providecommand{\href}[2]{#2}
\providecommand{\path}[1]{#1}
\providecommand{\DOIprefix}{doi:}
\providecommand{\ArXivprefix}{arXiv:}
\providecommand{\URLprefix}{URL: }
\providecommand{\Pubmedprefix}{pmid:}
\providecommand{\doi}[1]{\href{http://dx.doi.org/#1}{\path{#1}}}
\providecommand{\Pubmed}[1]{\href{pmid:#1}{\path{#1}}}
\providecommand{\bibinfo}[2]{#2}
\ifx\xfnm\relax \def\xfnm[#1]{\unskip,\space#1}\fi
\bibitem[{Bachute and Subhedar(2021)}]{bachute2021autonomous}
\bibinfo{author}{Bachute, M.R.}, \bibinfo{author}{Subhedar, J.M.}, \bibinfo{year}{2021}.
\newblock \bibinfo{title}{Autonomous driving architectures: insights of machine learning and deep learning algorithms}.
\newblock \bibinfo{journal}{Machine Learning with Applications} \bibinfo{volume}{6}, \bibinfo{pages}{100164}.
\bibitem[{Barcel{\'o} and Casas(2005)}]{barcelo2005dynamic}
\bibinfo{author}{Barcel{\'o}, J.}, \bibinfo{author}{Casas, J.}, \bibinfo{year}{2005}.
\newblock \bibinfo{title}{Dynamic network simulation with aimsun}, in: \bibinfo{booktitle}{Simulation approaches in transportation analysis: Recent advances and challenges}. \bibinfo{publisher}{Springer}, pp. \bibinfo{pages}{57--98}.
\bibitem[{Bergamini et~al.(2021)Bergamini, Ye, Scheel, Chen, Hu, Del~Pero, Osi{\'n}ski, Grimmett and Ondruska}]{bergamini2021simnet}
\bibinfo{author}{Bergamini, L.}, \bibinfo{author}{Ye, Y.}, \bibinfo{author}{Scheel, O.}, \bibinfo{author}{Chen, L.}, \bibinfo{author}{Hu, C.}, \bibinfo{author}{Del~Pero, L.}, \bibinfo{author}{Osi{\'n}ski, B.}, \bibinfo{author}{Grimmett, H.}, \bibinfo{author}{Ondruska, P.}, \bibinfo{year}{2021}.
\newblock \bibinfo{title}{{SimNet}: Learning reactive self-driving simulations from real-world observations}, in: \bibinfo{booktitle}{IEEE International Conference on Robotics and Automation (ICRA)}, \bibinfo{organization}{IEEE}. pp. \bibinfo{pages}{5119--5125}.
\bibitem[{Bockman et~al.(2024)Bockman, Howe, Orenstein and Dayoub}]{bockman2024aark}
\bibinfo{author}{Bockman, J.}, \bibinfo{author}{Howe, M.}, \bibinfo{author}{Orenstein, A.}, \bibinfo{author}{Dayoub, F.}, \bibinfo{year}{2024}.
\newblock \bibinfo{title}{Aark: An open toolkit for autonomous racing research}.
\newblock \bibinfo{journal}{arXiv preprint arXiv:2410.00358} .
\bibitem[{Bolte et~al.(2019)Bolte, Bar, Lipinski and Fingscheidt}]{bolte2019towards}
\bibinfo{author}{Bolte, J.A.}, \bibinfo{author}{Bar, A.}, \bibinfo{author}{Lipinski, D.}, \bibinfo{author}{Fingscheidt, T.}, \bibinfo{year}{2019}.
\newblock \bibinfo{title}{Towards corner case detection for autonomous driving}, in: \bibinfo{booktitle}{IEEE Intelligent vehicles symposium (IV)}, \bibinfo{organization}{IEEE}. pp. \bibinfo{pages}{438--445}.
\bibitem[{Browne et~al.(2012)Browne, Powley, Whitehouse, Lucas, Cowling, Rohlfshagen, Tavener, Perez, Samothrakis and Colton}]{browne2012survey}
\bibinfo{author}{Browne, C.B.}, \bibinfo{author}{Powley, E.}, \bibinfo{author}{Whitehouse, D.}, \bibinfo{author}{Lucas, S.M.}, \bibinfo{author}{Cowling, P.I.}, \bibinfo{author}{Rohlfshagen, P.}, \bibinfo{author}{Tavener, S.}, \bibinfo{author}{Perez, D.}, \bibinfo{author}{Samothrakis, S.}, \bibinfo{author}{Colton, S.}, \bibinfo{year}{2012}.
\newblock \bibinfo{title}{A survey of {Monte Carlo} tree search methods}.
\newblock \bibinfo{journal}{IEEE Transactions on Computational Intelligence and AI in Games} \bibinfo{volume}{4}, \bibinfo{pages}{1--43}.
\bibitem[{Caesar et~al.(2020)Caesar, Bankiti, Lang, Vora, Liong, Xu, Krishnan, Pan, Baldan and Beijbom}]{caesar2020nuscenes}
\bibinfo{author}{Caesar, H.}, \bibinfo{author}{Bankiti, V.}, \bibinfo{author}{Lang, A.H.}, \bibinfo{author}{Vora, S.}, \bibinfo{author}{Liong, V.E.}, \bibinfo{author}{Xu, Q.}, \bibinfo{author}{Krishnan, A.}, \bibinfo{author}{Pan, Y.}, \bibinfo{author}{Baldan, G.}, \bibinfo{author}{Beijbom, O.}, \bibinfo{year}{2020}.
\newblock \bibinfo{title}{{nuScenes}: A multimodal dataset for autonomous driving}, in: \bibinfo{booktitle}{IEEE/CVF Conference on Computer Vision and Pattern Recognition (CVPR)}, pp. \bibinfo{pages}{11621--11631}.
\bibitem[{Caesar et~al.(2021)Caesar, Kabzan, Tan, Fong, Wolff, Lang, Fletcher, Beijbom and Omari}]{caesar2021nuplan}
\bibinfo{author}{Caesar, H.}, \bibinfo{author}{Kabzan, J.}, \bibinfo{author}{Tan, K.S.}, \bibinfo{author}{Fong, W.K.}, \bibinfo{author}{Wolff, E.}, \bibinfo{author}{Lang, A.}, \bibinfo{author}{Fletcher, L.}, \bibinfo{author}{Beijbom, O.}, \bibinfo{author}{Omari, S.}, \bibinfo{year}{2021}.
\newblock \bibinfo{title}{{nuPlan}: A closed-loop {ML}-based planning benchmark for autonomous vehicles}.
\newblock \bibinfo{journal}{arXiv preprint arXiv:2106.11810} .
\bibitem[{Cameron and Duncan(1996)}]{cameron1996paramics}
\bibinfo{author}{Cameron, G.D.}, \bibinfo{author}{Duncan, G.I.}, \bibinfo{year}{1996}.
\newblock \bibinfo{title}{{PARAMICS}—parallel microscopic simulation of road traffic}.
\newblock \bibinfo{journal}{The Journal of Supercomputing} \bibinfo{volume}{10}, \bibinfo{pages}{25--53}.
\bibitem[{Carpin et~al.(2007)Carpin, Lewis, Wang, Balakirsky and Scrapper}]{carpin2007usarsim}
\bibinfo{author}{Carpin, S.}, \bibinfo{author}{Lewis, M.}, \bibinfo{author}{Wang, J.}, \bibinfo{author}{Balakirsky, S.}, \bibinfo{author}{Scrapper, C.}, \bibinfo{year}{2007}.
\newblock \bibinfo{title}{Usarsim: a robot simulator for research and education}, in: \bibinfo{booktitle}{IEEE International Conference on Robotics and Automation (ICRA)}, \bibinfo{organization}{IEEE}. pp. \bibinfo{pages}{1400--1405}.
\bibitem[{Casas et~al.(2021)Casas, Sadat and Urtasun}]{casas2021mp3}
\bibinfo{author}{Casas, S.}, \bibinfo{author}{Sadat, A.}, \bibinfo{author}{Urtasun, R.}, \bibinfo{year}{2021}.
\newblock \bibinfo{title}{{MP3}: A unified model to map, perceive, predict and plan}, in: \bibinfo{booktitle}{IEEE/CVF Conference on Computer Vision and Pattern Recognition (CVPR)}, pp. \bibinfo{pages}{14403--14412}.
\bibitem[{Chao et~al.(2020)Chao, Bi, Li, Mao, Wang, Lin and Deng}]{chao2020survey}
\bibinfo{author}{Chao, Q.}, \bibinfo{author}{Bi, H.}, \bibinfo{author}{Li, W.}, \bibinfo{author}{Mao, T.}, \bibinfo{author}{Wang, Z.}, \bibinfo{author}{Lin, M.C.}, \bibinfo{author}{Deng, Z.}, \bibinfo{year}{2020}.
\newblock \bibinfo{title}{A survey on visual traffic simulation: Models, evaluations, and applications in autonomous driving}, in: \bibinfo{booktitle}{Computer Graphics Forum}, \bibinfo{organization}{Wiley Online Library}. pp. \bibinfo{pages}{287--308}.
\bibitem[{Chen et~al.(2015)Chen, Seff, Kornhauser and Xiao}]{chen2015deepdriving}
\bibinfo{author}{Chen, C.}, \bibinfo{author}{Seff, A.}, \bibinfo{author}{Kornhauser, A.}, \bibinfo{author}{Xiao, J.}, \bibinfo{year}{2015}.
\newblock \bibinfo{title}{{DeepDriving}: Learning affordance for direct perception in autonomous driving}, in: \bibinfo{booktitle}{IEEE International Conference on Computer Vision (ICCV)}, pp. \bibinfo{pages}{2722--2730}.
\bibitem[{Chen et~al.(2024a)Chen, Sinavski, H{\"u}nermann, Karnsund, Willmott, Birch, Maund and Shotton}]{chen2024driving}
\bibinfo{author}{Chen, L.}, \bibinfo{author}{Sinavski, O.}, \bibinfo{author}{H{\"u}nermann, J.}, \bibinfo{author}{Karnsund, A.}, \bibinfo{author}{Willmott, A.J.}, \bibinfo{author}{Birch, D.}, \bibinfo{author}{Maund, D.}, \bibinfo{author}{Shotton, J.}, \bibinfo{year}{2024}a.
\newblock \bibinfo{title}{Driving with {LLMs}: Fusing object-level vector modality for explainable autonomous driving}, in: \bibinfo{booktitle}{IEEE International Conference on Robotics and Automation (ICRA)}, \bibinfo{organization}{IEEE}. pp. \bibinfo{pages}{14093--14100}.
\bibitem[{Chen et~al.(2024b)Chen, Wu, Chitta, Jaeger, Geiger and Li}]{chen2024end}
\bibinfo{author}{Chen, L.}, \bibinfo{author}{Wu, P.}, \bibinfo{author}{Chitta, K.}, \bibinfo{author}{Jaeger, B.}, \bibinfo{author}{Geiger, A.}, \bibinfo{author}{Li, H.}, \bibinfo{year}{2024}b.
\newblock \bibinfo{title}{End-to-end autonomous driving: Challenges and frontiers}.
\newblock \bibinfo{journal}{IEEE Transactions on Pattern Analysis and Machine Intelligence} .
\bibitem[{Chib and Singh(2023)}]{chib2023recent}
\bibinfo{author}{Chib, P.S.}, \bibinfo{author}{Singh, P.}, \bibinfo{year}{2023}.
\newblock \bibinfo{title}{Recent advancements in end-to-end autonomous driving using deep learning: A survey}.
\newblock \bibinfo{journal}{IEEE Transactions on Intelligent Vehicles} .
\bibitem[{Chitta et~al.(2021)Chitta, Prakash and Geiger}]{chitta2021neat}
\bibinfo{author}{Chitta, K.}, \bibinfo{author}{Prakash, A.}, \bibinfo{author}{Geiger, A.}, \bibinfo{year}{2021}.
\newblock \bibinfo{title}{{NEAT}: Neural attention fields for end-to-end autonomous driving}, in: \bibinfo{booktitle}{IEEE/CVF International Conference on Computer Vision (ICCV)}, pp. \bibinfo{pages}{15793--15803}.
\bibitem[{Chitta et~al.(2022)Chitta, Prakash, Jaeger, Yu, Renz and Geiger}]{chitta2022transfuser}
\bibinfo{author}{Chitta, K.}, \bibinfo{author}{Prakash, A.}, \bibinfo{author}{Jaeger, B.}, \bibinfo{author}{Yu, Z.}, \bibinfo{author}{Renz, K.}, \bibinfo{author}{Geiger, A.}, \bibinfo{year}{2022}.
\newblock \bibinfo{title}{Transfuser: Imitation with transformer-based sensor fusion for autonomous driving}.
\newblock \bibinfo{journal}{IEEE Transactions on Pattern Analysis and Machine Intelligence} \bibinfo{volume}{45}, \bibinfo{pages}{12878--12895}.
\bibitem[{Codevilla et~al.(2019)Codevilla, Santana, L{\'o}pez and Gaidon}]{codevilla2019exploring}
\bibinfo{author}{Codevilla, F.}, \bibinfo{author}{Santana, E.}, \bibinfo{author}{L{\'o}pez, A.M.}, \bibinfo{author}{Gaidon, A.}, \bibinfo{year}{2019}.
\newblock \bibinfo{title}{Exploring the limitations of behavior cloning for autonomous driving}, in: \bibinfo{booktitle}{IEEE/CVF International Conference on Computer Vision (ICCV)}, pp. \bibinfo{pages}{9329--9338}.
\bibitem[{Cui et~al.(2024a)Cui, Ma, Cao, Ye and Wang}]{cui2024drive}
\bibinfo{author}{Cui, C.}, \bibinfo{author}{Ma, Y.}, \bibinfo{author}{Cao, X.}, \bibinfo{author}{Ye, W.}, \bibinfo{author}{Wang, Z.}, \bibinfo{year}{2024}a.
\newblock \bibinfo{title}{Drive as you speak: Enabling human-like interaction with large language models in autonomous vehicles}, in: \bibinfo{booktitle}{IEEE/CVF Winter Conference on Applications of Computer Vision (WACV)}, pp. \bibinfo{pages}{902--909}.
\bibitem[{Cui et~al.(2024b)Cui, Ma, Cao, Ye, Zhou, Liang, Chen, Lu, Yang, Liao et~al.}]{cui2024survey}
\bibinfo{author}{Cui, C.}, \bibinfo{author}{Ma, Y.}, \bibinfo{author}{Cao, X.}, \bibinfo{author}{Ye, W.}, \bibinfo{author}{Zhou, Y.}, \bibinfo{author}{Liang, K.}, \bibinfo{author}{Chen, J.}, \bibinfo{author}{Lu, J.}, \bibinfo{author}{Yang, Z.}, \bibinfo{author}{Liao, K.D.}, et~al., \bibinfo{year}{2024}b.
\newblock \bibinfo{title}{A survey on multimodal large language models for autonomous driving}, in: \bibinfo{booktitle}{IEEE/CVF Winter Conference on Applications of Computer Vision (WACV)}, pp. \bibinfo{pages}{958--979}.
\bibitem[{Cui et~al.(2023)Cui, Huang, Zhong, Liu, Wang, Sun, Li, Wang and Khajepour}]{cui2023drivellm}
\bibinfo{author}{Cui, Y.}, \bibinfo{author}{Huang, S.}, \bibinfo{author}{Zhong, J.}, \bibinfo{author}{Liu, Z.}, \bibinfo{author}{Wang, Y.}, \bibinfo{author}{Sun, C.}, \bibinfo{author}{Li, B.}, \bibinfo{author}{Wang, X.}, \bibinfo{author}{Khajepour, A.}, \bibinfo{year}{2023}.
\newblock \bibinfo{title}{{DriveLLM}: Charting the path toward full autonomous driving with large language models}.
\newblock \bibinfo{journal}{IEEE Transactions on Intelligent Vehicles} .
\bibitem[{Dauner et~al.(2023)Dauner, Hallgarten, Geiger and Chitta}]{Dauner2023CORL}
\bibinfo{author}{Dauner, D.}, \bibinfo{author}{Hallgarten, M.}, \bibinfo{author}{Geiger, A.}, \bibinfo{author}{Chitta, K.}, \bibinfo{year}{2023}.
\newblock \bibinfo{title}{Parting with misconceptions about learning-based vehicle motion planning}, in: \bibinfo{booktitle}{Conference on Robot Learning (CoRL)}, pp. \bibinfo{pages}{1268--1281}.
\bibitem[{Dosovitskiy et~al.(2017)Dosovitskiy, Ros, Codevilla, Lopez and Koltun}]{dosovitskiy2017carla}
\bibinfo{author}{Dosovitskiy, A.}, \bibinfo{author}{Ros, G.}, \bibinfo{author}{Codevilla, F.}, \bibinfo{author}{Lopez, A.}, \bibinfo{author}{Koltun, V.}, \bibinfo{year}{2017}.
\newblock \bibinfo{title}{{CARLA}: An open urban driving simulator}, in: \bibinfo{booktitle}{Conference on Robot Learning (CoRL)}, \bibinfo{organization}{PMLR}. pp. \bibinfo{pages}{1--16}.
\bibitem[{Fellendorf and Vortisch(2010)}]{fellendorf2010microscopic}
\bibinfo{author}{Fellendorf, M.}, \bibinfo{author}{Vortisch, P.}, \bibinfo{year}{2010}.
\newblock \bibinfo{title}{Microscopic traffic flow simulator {VISSIM}}.
\newblock \bibinfo{journal}{Fundamentals of Traffic Simulation} , \bibinfo{pages}{63--93}.
\bibitem[{Feng et~al.(2020)Feng, Haase-Sch{\"u}tz, Rosenbaum, Hertlein, Glaeser, Timm, Wiesbeck and Dietmayer}]{feng2020deep}
\bibinfo{author}{Feng, D.}, \bibinfo{author}{Haase-Sch{\"u}tz, C.}, \bibinfo{author}{Rosenbaum, L.}, \bibinfo{author}{Hertlein, H.}, \bibinfo{author}{Glaeser, C.}, \bibinfo{author}{Timm, F.}, \bibinfo{author}{Wiesbeck, W.}, \bibinfo{author}{Dietmayer, K.}, \bibinfo{year}{2020}.
\newblock \bibinfo{title}{Deep multi-modal object detection and semantic segmentation for autonomous driving: Datasets, methods, and challenges}.
\newblock \bibinfo{journal}{IEEE Transactions on Intelligent Transportation Systems} \bibinfo{volume}{22}, \bibinfo{pages}{1341--1360}.
\bibitem[{Feng et~al.(2023)Feng, Li, Peng, Tan and Zhou}]{feng2023trafficgen}
\bibinfo{author}{Feng, L.}, \bibinfo{author}{Li, Q.}, \bibinfo{author}{Peng, Z.}, \bibinfo{author}{Tan, S.}, \bibinfo{author}{Zhou, B.}, \bibinfo{year}{2023}.
\newblock \bibinfo{title}{{TrafficGen}: Learning to generate diverse and realistic traffic scenarios}, in: \bibinfo{booktitle}{IEEE International Conference on Robotics and Automation (ICRA)}, \bibinfo{organization}{IEEE}. pp. \bibinfo{pages}{3567--3575}.
\bibitem[{Fu et~al.(2024)Fu, Lei, Wen, Cai, Mao, Dou, Shi and Qiao}]{fu2024limsim++}
\bibinfo{author}{Fu, D.}, \bibinfo{author}{Lei, W.}, \bibinfo{author}{Wen, L.}, \bibinfo{author}{Cai, P.}, \bibinfo{author}{Mao, S.}, \bibinfo{author}{Dou, M.}, \bibinfo{author}{Shi, B.}, \bibinfo{author}{Qiao, Y.}, \bibinfo{year}{2024}.
\newblock \bibinfo{title}{{LimSim++}: A closed-loop platform for deploying multimodal {LLMs} in autonomous driving}.
\newblock \bibinfo{journal}{arXiv preprint arXiv:2402.01246} .
\bibitem[{Geiger et~al.(2013)Geiger, Lenz, Stiller and Urtasun}]{geiger2013vision}
\bibinfo{author}{Geiger, A.}, \bibinfo{author}{Lenz, P.}, \bibinfo{author}{Stiller, C.}, \bibinfo{author}{Urtasun, R.}, \bibinfo{year}{2013}.
\newblock \bibinfo{title}{Vision meets robotics: The {KITTI} dataset}.
\newblock \bibinfo{journal}{The International Journal of Robotics Research} \bibinfo{volume}{32}, \bibinfo{pages}{1231--1237}.
\bibitem[{Gog et~al.(2021)Gog, Kalra, Schafhalter, Wright, Gonzalez and Stoica}]{gog2021pylot}
\bibinfo{author}{Gog, I.}, \bibinfo{author}{Kalra, S.}, \bibinfo{author}{Schafhalter, P.}, \bibinfo{author}{Wright, M.A.}, \bibinfo{author}{Gonzalez, J.E.}, \bibinfo{author}{Stoica, I.}, \bibinfo{year}{2021}.
\newblock \bibinfo{title}{Pylot: A modular platform for exploring latency-accuracy tradeoffs in autonomous vehicles}, in: \bibinfo{booktitle}{IEEE International Conference on Robotics and Automation (ICRA)}, \bibinfo{organization}{IEEE}. pp. \bibinfo{pages}{8806--8813}.
\bibitem[{Grigorescu et~al.(2020)Grigorescu, Trasnea, Cocias and Macesanu}]{grigorescu2020survey}
\bibinfo{author}{Grigorescu, S.}, \bibinfo{author}{Trasnea, B.}, \bibinfo{author}{Cocias, T.}, \bibinfo{author}{Macesanu, G.}, \bibinfo{year}{2020}.
\newblock \bibinfo{title}{A survey of deep learning techniques for autonomous driving}.
\newblock \bibinfo{journal}{Journal of Field Robotics} \bibinfo{volume}{37}, \bibinfo{pages}{362--386}.
\bibitem[{Gulino et~al.(2024)Gulino, Fu, Luo, Tucker, Bronstein, Lu, Harb, Pan, Wang, Chen et~al.}]{gulino2024waymax}
\bibinfo{author}{Gulino, C.}, \bibinfo{author}{Fu, J.}, \bibinfo{author}{Luo, W.}, \bibinfo{author}{Tucker, G.}, \bibinfo{author}{Bronstein, E.}, \bibinfo{author}{Lu, Y.}, \bibinfo{author}{Harb, J.}, \bibinfo{author}{Pan, X.}, \bibinfo{author}{Wang, Y.}, \bibinfo{author}{Chen, X.}, et~al., \bibinfo{year}{2024}.
\newblock \bibinfo{title}{Waymax: An accelerated, data-driven simulator for large-scale autonomous driving research}.
\newblock \bibinfo{journal}{Advances in Neural Information Processing Systems (NeurIPS)} \bibinfo{volume}{36}.
\bibitem[{Guo et~al.(2019)Guo, Kurup and Shah}]{guo2019safe}
\bibinfo{author}{Guo, J.}, \bibinfo{author}{Kurup, U.}, \bibinfo{author}{Shah, M.}, \bibinfo{year}{2019}.
\newblock \bibinfo{title}{Is it safe to drive? an overview of factors, metrics, and datasets for driveability assessment in autonomous driving}.
\newblock \bibinfo{journal}{IEEE Transactions on Intelligent Transportation Systems} \bibinfo{volume}{21}, \bibinfo{pages}{3135--3151}.
\bibitem[{Haas(2014)}]{haas2014history}
\bibinfo{author}{Haas, J.K.}, \bibinfo{year}{2014}.
\newblock \bibinfo{title}{A history of the unity game engine}.
\bibitem[{Halati et~al.(1997)Halati, Lieu and Walker}]{halati1997corsim}
\bibinfo{author}{Halati, A.}, \bibinfo{author}{Lieu, H.}, \bibinfo{author}{Walker, S.}, \bibinfo{year}{1997}.
\newblock \bibinfo{title}{{CORSIM}-corridor traffic simulation model}, in: \bibinfo{booktitle}{Traffic Congestion and Traffic Safety in the 21st Century: Challenges, Innovations, and Opportunities}.
\bibitem[{Hallyburton et~al.(2023)Hallyburton, Zhang and Pajic}]{hallyburton2023avstack}
\bibinfo{author}{Hallyburton, R.S.}, \bibinfo{author}{Zhang, S.}, \bibinfo{author}{Pajic, M.}, \bibinfo{year}{2023}.
\newblock \bibinfo{title}{Avstack: An open-source, reconfigurable platform for autonomous vehicle development}, in: \bibinfo{booktitle}{International Conference on Cyber-Physical Systems (with CPS-IoT Week 2023)}, pp. \bibinfo{pages}{209--220}.
\bibitem[{Hu et~al.(2023a)Hu, Li, Huang, Tang, Huai and Chen}]{hu2023simulation}
\bibinfo{author}{Hu, X.}, \bibinfo{author}{Li, S.}, \bibinfo{author}{Huang, T.}, \bibinfo{author}{Tang, B.}, \bibinfo{author}{Huai, R.}, \bibinfo{author}{Chen, L.}, \bibinfo{year}{2023}a.
\newblock \bibinfo{title}{How simulation helps autonomous driving: A survey of sim2real, digital twins, and parallel intelligence}.
\newblock \bibinfo{journal}{IEEE Transactions on Intelligent Vehicles} .
\bibitem[{Hu et~al.(2023b)Hu, Yang, Chen, Li, Sima, Zhu, Chai, Du, Lin, Wang et~al.}]{hu2023planning}
\bibinfo{author}{Hu, Y.}, \bibinfo{author}{Yang, J.}, \bibinfo{author}{Chen, L.}, \bibinfo{author}{Li, K.}, \bibinfo{author}{Sima, C.}, \bibinfo{author}{Zhu, X.}, \bibinfo{author}{Chai, S.}, \bibinfo{author}{Du, S.}, \bibinfo{author}{Lin, T.}, \bibinfo{author}{Wang, W.}, et~al., \bibinfo{year}{2023}b.
\newblock \bibinfo{title}{Planning-oriented autonomous driving}, in: \bibinfo{booktitle}{IEEE/CVF Conference on Computer Vision and Pattern Recognition (CVPR)}, pp. \bibinfo{pages}{17853--17862}.
\bibitem[{Huang and Chen(2020)}]{huang2020autonomous}
\bibinfo{author}{Huang, Y.}, \bibinfo{author}{Chen, Y.}, \bibinfo{year}{2020}.
\newblock \bibinfo{title}{Autonomous driving with deep learning: A survey of state-of-art technologies}.
\newblock \bibinfo{journal}{arXiv preprint arXiv:2006.06091} .
\bibitem[{Huang et~al.(2022)Huang, Du, Yang, Zhou, Zhang and Chen}]{huang2022survey}
\bibinfo{author}{Huang, Y.}, \bibinfo{author}{Du, J.}, \bibinfo{author}{Yang, Z.}, \bibinfo{author}{Zhou, Z.}, \bibinfo{author}{Zhang, L.}, \bibinfo{author}{Chen, H.}, \bibinfo{year}{2022}.
\newblock \bibinfo{title}{A survey on trajectory-prediction methods for autonomous driving}.
\newblock \bibinfo{journal}{IEEE Transactions on Intelligent Vehicles} \bibinfo{volume}{7}, \bibinfo{pages}{652--674}.
\bibitem[{Jia et~al.(2024)Jia, Yang, Li, Zhang and Yan}]{jia2024bench2drive}
\bibinfo{author}{Jia, X.}, \bibinfo{author}{Yang, Z.}, \bibinfo{author}{Li, Q.}, \bibinfo{author}{Zhang, Z.}, \bibinfo{author}{Yan, J.}, \bibinfo{year}{2024}.
\newblock \bibinfo{title}{{Bench2Drive}: Towards multi-ability benchmarking of closed-loop end-to-end autonomous driving}.
\newblock \bibinfo{journal}{arXiv preprint arXiv:2406.03877} .
\bibitem[{Jin et~al.(2023)Jin, Shen, Peng, Liu, Qin, Li, Xie, Gao, Zhou and Gong}]{jin2023surrealdriver}
\bibinfo{author}{Jin, Y.}, \bibinfo{author}{Shen, X.}, \bibinfo{author}{Peng, H.}, \bibinfo{author}{Liu, X.}, \bibinfo{author}{Qin, J.}, \bibinfo{author}{Li, J.}, \bibinfo{author}{Xie, J.}, \bibinfo{author}{Gao, P.}, \bibinfo{author}{Zhou, G.}, \bibinfo{author}{Gong, J.}, \bibinfo{year}{2023}.
\newblock \bibinfo{title}{{SurrealDriver}: Designing generative driver agent simulation framework in urban contexts based on large language model}.
\newblock \bibinfo{journal}{arXiv preprint arXiv:2309.13193} .
\bibitem[{Kesting et~al.(2007)Kesting, Treiber and Helbing}]{kesting2007general}
\bibinfo{author}{Kesting, A.}, \bibinfo{author}{Treiber, M.}, \bibinfo{author}{Helbing, D.}, \bibinfo{year}{2007}.
\newblock \bibinfo{title}{General lane-changing model {MOBIL} for car-following models}.
\newblock \bibinfo{journal}{Transportation Research Record} \bibinfo{volume}{1999}, \bibinfo{pages}{86--94}.
\bibitem[{Kiran et~al.(2021)Kiran, Sobh, Talpaert, Mannion, Al~Sallab, Yogamani and P{\'e}rez}]{kiran2021deep}
\bibinfo{author}{Kiran, B.R.}, \bibinfo{author}{Sobh, I.}, \bibinfo{author}{Talpaert, V.}, \bibinfo{author}{Mannion, P.}, \bibinfo{author}{Al~Sallab, A.A.}, \bibinfo{author}{Yogamani, S.}, \bibinfo{author}{P{\'e}rez, P.}, \bibinfo{year}{2021}.
\newblock \bibinfo{title}{Deep reinforcement learning for autonomous driving: A survey}.
\newblock \bibinfo{journal}{IEEE Transactions on Intelligent Transportation Systems} \bibinfo{volume}{23}, \bibinfo{pages}{4909--4926}.
\bibitem[{Koenig and Howard(2004)}]{koenig2004design}
\bibinfo{author}{Koenig, N.}, \bibinfo{author}{Howard, A.}, \bibinfo{year}{2004}.
\newblock \bibinfo{title}{Design and use paradigms for gazebo, an open-source multi-robot simulator}, in: \bibinfo{booktitle}{IEEE/RSJ international conference on intelligent robots and systems (IROS)(IEEE Cat. No. 04CH37566)}, \bibinfo{organization}{Ieee}. pp. \bibinfo{pages}{2149--2154}.
\bibitem[{Kotusevski and Hawick(2009)}]{kotusevski2009review}
\bibinfo{author}{Kotusevski, G.}, \bibinfo{author}{Hawick, K.A.}, \bibinfo{year}{2009}.
\newblock \bibinfo{title}{A review of traffic simulation software}.
\newblock \bibinfo{journal}{Research Letters in the Information and Mathematical Sciences} \bibinfo{volume}{13}, \bibinfo{pages}{35--54}.
\bibitem[{Lan and Tian(2022)}]{lan2022instance}
\bibinfo{author}{Lan, Q.}, \bibinfo{author}{Tian, Q.}, \bibinfo{year}{2022}.
\newblock \bibinfo{title}{Instance, scale, and teacher adaptive knowledge distillation for visual detection in autonomous driving}.
\newblock \bibinfo{journal}{IEEE Transactions on Intelligent Vehicles} \bibinfo{volume}{8}, \bibinfo{pages}{2358--2370}.
\bibitem[{Le~Mero et~al.(2022)Le~Mero, Yi, Dianati and Mouzakitis}]{le2022survey}
\bibinfo{author}{Le~Mero, L.}, \bibinfo{author}{Yi, D.}, \bibinfo{author}{Dianati, M.}, \bibinfo{author}{Mouzakitis, A.}, \bibinfo{year}{2022}.
\newblock \bibinfo{title}{A survey on imitation learning techniques for end-to-end autonomous vehicles}.
\newblock \bibinfo{journal}{IEEE Transactions on Intelligent Transportation Systems} \bibinfo{volume}{23}, \bibinfo{pages}{14128--14147}.
\bibitem[{Li et~al.(2022)Li, Chen, Wang, Hong, Ye, Han, Chen, Zhang, Xu, Yeung et~al.}]{li2022coda}
\bibinfo{author}{Li, K.}, \bibinfo{author}{Chen, K.}, \bibinfo{author}{Wang, H.}, \bibinfo{author}{Hong, L.}, \bibinfo{author}{Ye, C.}, \bibinfo{author}{Han, J.}, \bibinfo{author}{Chen, Y.}, \bibinfo{author}{Zhang, W.}, \bibinfo{author}{Xu, C.}, \bibinfo{author}{Yeung, D.Y.}, et~al., \bibinfo{year}{2022}.
\newblock \bibinfo{title}{{CODA}: A real-world road corner case dataset for object detection in autonomous driving}, in: \bibinfo{booktitle}{European Conference on Computer Vision (ECCV)}, \bibinfo{organization}{Springer}. pp. \bibinfo{pages}{406--423}.
\bibitem[{Li et~al.(2024a)Li, Shao, Dong, Tian, Zhang, Yang and Zhang}]{li2024data}
\bibinfo{author}{Li, L.}, \bibinfo{author}{Shao, W.}, \bibinfo{author}{Dong, W.}, \bibinfo{author}{Tian, Y.}, \bibinfo{author}{Zhang, Q.}, \bibinfo{author}{Yang, K.}, \bibinfo{author}{Zhang, W.}, \bibinfo{year}{2024}a.
\newblock \bibinfo{title}{Data-centric evolution in autonomous driving: A comprehensive survey of big data system, data mining, and closed-loop technologies}.
\newblock \bibinfo{journal}{arXiv preprint arXiv:2401.12888} .
\bibitem[{Li et~al.(2023a)Li, Bai, Cai, Wen, Fu, Zhang, Yang, Cai, Ma, Guo et~al.}]{li2023towards}
\bibinfo{author}{Li, X.}, \bibinfo{author}{Bai, Y.}, \bibinfo{author}{Cai, P.}, \bibinfo{author}{Wen, L.}, \bibinfo{author}{Fu, D.}, \bibinfo{author}{Zhang, B.}, \bibinfo{author}{Yang, X.}, \bibinfo{author}{Cai, X.}, \bibinfo{author}{Ma, T.}, \bibinfo{author}{Guo, J.}, et~al., \bibinfo{year}{2023}a.
\newblock \bibinfo{title}{Towards knowledge-driven autonomous driving}.
\newblock \bibinfo{journal}{arXiv preprint arXiv:2312.04316} .
\bibitem[{Li et~al.(2023b)Li, Wang, Huang and Chen}]{li2023survey}
\bibinfo{author}{Li, X.}, \bibinfo{author}{Wang, Z.}, \bibinfo{author}{Huang, Y.}, \bibinfo{author}{Chen, H.}, \bibinfo{year}{2023}b.
\newblock \bibinfo{title}{A survey on self-evolving autonomous driving: a perspective on data closed-loop technology}.
\newblock \bibinfo{journal}{IEEE Transactions on Intelligent Vehicles} .
\bibitem[{Li et~al.(2024b)Li, Yuan, Zhang, Yan, Shen, Wang and Yang}]{li2024choose}
\bibinfo{author}{Li, Y.}, \bibinfo{author}{Yuan, W.}, \bibinfo{author}{Zhang, S.}, \bibinfo{author}{Yan, W.}, \bibinfo{author}{Shen, Q.}, \bibinfo{author}{Wang, C.}, \bibinfo{author}{Yang, M.}, \bibinfo{year}{2024}b.
\newblock \bibinfo{title}{Choose your simulator wisely: A review on open-source simulators for autonomous driving}.
\newblock \bibinfo{journal}{IEEE Transactions on Intelligent Vehicles} .
\bibitem[{Li et~al.(2024c)Li, Yu, Lan, Li, Kautz, Lu and Alvarez}]{li2024ego}
\bibinfo{author}{Li, Z.}, \bibinfo{author}{Yu, Z.}, \bibinfo{author}{Lan, S.}, \bibinfo{author}{Li, J.}, \bibinfo{author}{Kautz, J.}, \bibinfo{author}{Lu, T.}, \bibinfo{author}{Alvarez, J.M.}, \bibinfo{year}{2024}c.
\newblock \bibinfo{title}{Is ego status all you need for open-loop end-to-end autonomous driving?}, in: \bibinfo{booktitle}{IEEE/CVF Conference on Computer Vision and Pattern Recognition (CVPR)}, pp. \bibinfo{pages}{14864--14873}.
\bibitem[{Liu et~al.(2024)Liu, Yurtsever, Fossaert, Zhou, Zimmer, Cui, Zagar and Knoll}]{liu2024survey}
\bibinfo{author}{Liu, M.}, \bibinfo{author}{Yurtsever, E.}, \bibinfo{author}{Fossaert, J.}, \bibinfo{author}{Zhou, X.}, \bibinfo{author}{Zimmer, W.}, \bibinfo{author}{Cui, Y.}, \bibinfo{author}{Zagar, B.L.}, \bibinfo{author}{Knoll, A.C.}, \bibinfo{year}{2024}.
\newblock \bibinfo{title}{A survey on autonomous driving datasets: Statistics, annotation quality, and a future outlook}.
\newblock \bibinfo{journal}{IEEE Transactions on Intelligent Vehicles} .
\bibitem[{Ljungbergh et~al.(2025)Ljungbergh, Tonderski, Johnander, Caesar, {\AA}str{\"o}m, Felsberg and Petersson}]{ljungbergh2025neuroncap}
\bibinfo{author}{Ljungbergh, W.}, \bibinfo{author}{Tonderski, A.}, \bibinfo{author}{Johnander, J.}, \bibinfo{author}{Caesar, H.}, \bibinfo{author}{{\AA}str{\"o}m, K.}, \bibinfo{author}{Felsberg, M.}, \bibinfo{author}{Petersson, C.}, \bibinfo{year}{2025}.
\newblock \bibinfo{title}{{NeuroNCAP}: Photorealistic closed-loop safety testing for autonomous driving}, in: \bibinfo{booktitle}{European Conference on Computer Vision (ECCV)}, \bibinfo{organization}{Springer}. pp. \bibinfo{pages}{161--177}.
\bibitem[{Maddern et~al.(2017)Maddern, Pascoe, Linegar and Newman}]{maddern20171}
\bibinfo{author}{Maddern, W.}, \bibinfo{author}{Pascoe, G.}, \bibinfo{author}{Linegar, C.}, \bibinfo{author}{Newman, P.}, \bibinfo{year}{2017}.
\newblock \bibinfo{title}{1 year, 1000 km: The {Oxford} {RobotCar} dataset}.
\newblock \bibinfo{journal}{The International Journal of Robotics Research} \bibinfo{volume}{36}, \bibinfo{pages}{3--15}.
\bibitem[{Mao et~al.(2023)Mao, Qian, Ye, Zhao and Wang}]{mao2023gpt}
\bibinfo{author}{Mao, J.}, \bibinfo{author}{Qian, Y.}, \bibinfo{author}{Ye, J.}, \bibinfo{author}{Zhao, H.}, \bibinfo{author}{Wang, Y.}, \bibinfo{year}{2023}.
\newblock \bibinfo{title}{Gpt-driver: Learning to drive with {GPT}}.
\newblock \bibinfo{journal}{arXiv preprint arXiv:2310.01415} .
\bibitem[{Mei et~al.(2024)Mei, Ma, Yang, Wen, Cai, Li, Fu, Zhang, Cai, Dou et~al.}]{mei2024continuously}
\bibinfo{author}{Mei, J.}, \bibinfo{author}{Ma, Y.}, \bibinfo{author}{Yang, X.}, \bibinfo{author}{Wen, L.}, \bibinfo{author}{Cai, X.}, \bibinfo{author}{Li, X.}, \bibinfo{author}{Fu, D.}, \bibinfo{author}{Zhang, B.}, \bibinfo{author}{Cai, P.}, \bibinfo{author}{Dou, M.}, et~al., \bibinfo{year}{2024}.
\newblock \bibinfo{title}{Continuously learning, adapting, and improving: A dual-process approach to autonomous driving}.
\newblock \bibinfo{journal}{arXiv preprint arXiv:2405.15324} \bibinfo{note}{\nolinebreak.}
\bibitem[{Michel(2004)}]{michel2004cyberbotics}
\bibinfo{author}{Michel, O.}, \bibinfo{year}{2004}.
\newblock \bibinfo{title}{Cyberbotics ltd. webots™: professional mobile robot simulation}.
\newblock \bibinfo{journal}{International Journal of Advanced Robotic Systems} \bibinfo{volume}{1}, \bibinfo{pages}{5}.
\bibitem[{Mozaffari et~al.(2020)Mozaffari, Al-Jarrah, Dianati, Jennings and Mouzakitis}]{mozaffari2020deep}
\bibinfo{author}{Mozaffari, S.}, \bibinfo{author}{Al-Jarrah, O.Y.}, \bibinfo{author}{Dianati, M.}, \bibinfo{author}{Jennings, P.}, \bibinfo{author}{Mouzakitis, A.}, \bibinfo{year}{2020}.
\newblock \bibinfo{title}{Deep learning-based vehicle behavior prediction for autonomous driving applications: A review}.
\newblock \bibinfo{journal}{IEEE Transactions on Intelligent Transportation Systems} \bibinfo{volume}{23}, \bibinfo{pages}{33--47}.
\bibitem[{M{\"u}tsch et~al.(2023)M{\"u}tsch, Gremmelmaier, Becker, Bogdoll, Zofka and Z{\"o}llner}]{mutsch2023model}
\bibinfo{author}{M{\"u}tsch, F.}, \bibinfo{author}{Gremmelmaier, H.}, \bibinfo{author}{Becker, N.}, \bibinfo{author}{Bogdoll, D.}, \bibinfo{author}{Zofka, M.R.}, \bibinfo{author}{Z{\"o}llner, J.M.}, \bibinfo{year}{2023}.
\newblock \bibinfo{title}{From model-based to data-driven simulation: Challenges and trends in autonomous driving}.
\newblock \bibinfo{journal}{arXiv preprint arXiv:2305.13960} .
\bibitem[{Ochs et~al.(2024)Ochs, Doll, Grimm, Fleck, Heinrich, Orf, Schotschneider, Gremmelmaier, Polley, Pavlitska et~al.}]{ochs2024one}
\bibinfo{author}{Ochs, S.}, \bibinfo{author}{Doll, J.}, \bibinfo{author}{Grimm, D.}, \bibinfo{author}{Fleck, T.}, \bibinfo{author}{Heinrich, M.}, \bibinfo{author}{Orf, S.}, \bibinfo{author}{Schotschneider, A.}, \bibinfo{author}{Gremmelmaier, H.}, \bibinfo{author}{Polley, R.}, \bibinfo{author}{Pavlitska, S.}, et~al., \bibinfo{year}{2024}.
\newblock \bibinfo{title}{One stack to rule them all: To drive automated vehicles, and reach for the 4th level}.
\newblock \bibinfo{journal}{arXiv preprint arXiv:2404.02645} .
\bibitem[{Pomerleau(1988)}]{pomerleau1988alvinn}
\bibinfo{author}{Pomerleau, D.A.}, \bibinfo{year}{1988}.
\newblock \bibinfo{title}{{ALVINN}: An autonomous land vehicle in a neural network}.
\newblock \bibinfo{journal}{Advances in neural information processing systems} \bibinfo{volume}{1}.
\bibitem[{Quigley et~al.(2009)Quigley, Conley, Gerkey, Faust, Foote, Leibs, Wheeler, Ng et~al.}]{quigley2009ros}
\bibinfo{author}{Quigley, M.}, \bibinfo{author}{Conley, K.}, \bibinfo{author}{Gerkey, B.}, \bibinfo{author}{Faust, J.}, \bibinfo{author}{Foote, T.}, \bibinfo{author}{Leibs, J.}, \bibinfo{author}{Wheeler, R.}, \bibinfo{author}{Ng, A.Y.}, et~al., \bibinfo{year}{2009}.
\newblock \bibinfo{title}{Ros: an open-source robot operating system}, in: \bibinfo{booktitle}{ICRA workshop on open source software}, \bibinfo{organization}{Kobe, Japan}. p.~\bibinfo{pages}{5}.
\bibitem[{Ramanishka et~al.(2018)Ramanishka, Chen, Misu and Saenko}]{ramanishka2018toward}
\bibinfo{author}{Ramanishka, V.}, \bibinfo{author}{Chen, Y.T.}, \bibinfo{author}{Misu, T.}, \bibinfo{author}{Saenko, K.}, \bibinfo{year}{2018}.
\newblock \bibinfo{title}{Toward driving scene understanding: A dataset for learning driver behavior and causal reasoning}, in: \bibinfo{booktitle}{IEEE Conference on Computer Vision and Pattern Recognition (CVPR)}, pp. \bibinfo{pages}{7699--7707}.
\bibitem[{Rong et~al.(2020)Rong, Shin, Tabatabaee, Lu, Lemke, Mo{\v{z}}eiko, Boise, Uhm, Gerow, Mehta et~al.}]{rong2020lgsvl}
\bibinfo{author}{Rong, G.}, \bibinfo{author}{Shin, B.H.}, \bibinfo{author}{Tabatabaee, H.}, \bibinfo{author}{Lu, Q.}, \bibinfo{author}{Lemke, S.}, \bibinfo{author}{Mo{\v{z}}eiko, M.}, \bibinfo{author}{Boise, E.}, \bibinfo{author}{Uhm, G.}, \bibinfo{author}{Gerow, M.}, \bibinfo{author}{Mehta, S.}, et~al., \bibinfo{year}{2020}.
\newblock \bibinfo{title}{Lgsvl simulator: A high fidelity simulator for autonomous driving}, in: \bibinfo{booktitle}{IEEE International conference on intelligent transportation systems (ITSC)}, \bibinfo{organization}{IEEE}. pp. \bibinfo{pages}{1--6}.
\bibitem[{Sanders(2016)}]{sanders2016introduction}
\bibinfo{author}{Sanders, A.}, \bibinfo{year}{2016}.
\newblock \bibinfo{title}{An introduction to {Unreal Engine} 4}.
\bibitem[{Sha et~al.(2023)Sha, Mu, Jiang, Chen, Xu, Luo, Li, Tomizuka, Zhan and Ding}]{sha2023languagempc}
\bibinfo{author}{Sha, H.}, \bibinfo{author}{Mu, Y.}, \bibinfo{author}{Jiang, Y.}, \bibinfo{author}{Chen, L.}, \bibinfo{author}{Xu, C.}, \bibinfo{author}{Luo, P.}, \bibinfo{author}{Li, S.E.}, \bibinfo{author}{Tomizuka, M.}, \bibinfo{author}{Zhan, W.}, \bibinfo{author}{Ding, M.}, \bibinfo{year}{2023}.
\newblock \bibinfo{title}{{LanguageMPC}: Large language models as decision makers for autonomous driving}.
\newblock \bibinfo{journal}{arXiv preprint arXiv:2310.03026} .
\bibitem[{Shah et~al.(2018)Shah, Dey, Lovett and Kapoor}]{shah2018airsim}
\bibinfo{author}{Shah, S.}, \bibinfo{author}{Dey, D.}, \bibinfo{author}{Lovett, C.}, \bibinfo{author}{Kapoor, A.}, \bibinfo{year}{2018}.
\newblock \bibinfo{title}{{AirSim}: High-fidelity visual and physical simulation for autonomous vehicles}, in: \bibinfo{booktitle}{Field and Service Robotics: Results of the 11th International Conference}, \bibinfo{organization}{Springer}. pp. \bibinfo{pages}{621--635}.
\bibitem[{Shao et~al.(2022)Shao, Wang, Chen, Li and Liu}]{shao2022interfuser}
\bibinfo{author}{Shao, H.}, \bibinfo{author}{Wang, L.}, \bibinfo{author}{Chen, R.}, \bibinfo{author}{Li, H.}, \bibinfo{author}{Liu, Y.}, \bibinfo{year}{2022}.
\newblock \bibinfo{title}{Safety-enhanced autonomous driving using interpretable sensor fusion transformer}.
\newblock \bibinfo{journal}{arXiv preprint arXiv:2207.14024} .
\bibitem[{Shao et~al.(2023)Shao, Wang, Chen, Li and Liu}]{shao2023safety}
\bibinfo{author}{Shao, H.}, \bibinfo{author}{Wang, L.}, \bibinfo{author}{Chen, R.}, \bibinfo{author}{Li, H.}, \bibinfo{author}{Liu, Y.}, \bibinfo{year}{2023}.
\newblock \bibinfo{title}{Safety-enhanced autonomous driving using interpretable sensor fusion transformer}, in: \bibinfo{booktitle}{Conference on Robot Learning (CoRL)}, \bibinfo{organization}{PMLR}. pp. \bibinfo{pages}{726--737}.
\bibitem[{Sun et~al.(2020)Sun, Kretzschmar, Dotiwalla, Chouard, Patnaik, Tsui, Guo, Zhou, Chai, Caine et~al.}]{sun2020scalability}
\bibinfo{author}{Sun, P.}, \bibinfo{author}{Kretzschmar, H.}, \bibinfo{author}{Dotiwalla, X.}, \bibinfo{author}{Chouard, A.}, \bibinfo{author}{Patnaik, V.}, \bibinfo{author}{Tsui, P.}, \bibinfo{author}{Guo, J.}, \bibinfo{author}{Zhou, Y.}, \bibinfo{author}{Chai, Y.}, \bibinfo{author}{Caine, B.}, et~al., \bibinfo{year}{2020}.
\newblock \bibinfo{title}{Scalability in perception for autonomous driving: Waymo open dataset}, in: \bibinfo{booktitle}{IEEE/CVF Conference on Computer Vision and Pattern Recognition (CVPR)}, pp. \bibinfo{pages}{2446--2454}.
\bibitem[{Sun et~al.(2022)Sun, Huang, Williams and Zhao}]{sun2022intersim}
\bibinfo{author}{Sun, Q.}, \bibinfo{author}{Huang, X.}, \bibinfo{author}{Williams, B.C.}, \bibinfo{author}{Zhao, H.}, \bibinfo{year}{2022}.
\newblock \bibinfo{title}{Intersim: Interactive traffic simulation via explicit relation modeling}, in: \bibinfo{booktitle}{IEEE/RSJ International Conference on Intelligent Robots and Systems (IROS)}, \bibinfo{organization}{IEEE}. pp. \bibinfo{pages}{11416--11423}.
\bibitem[{Tampuu et~al.(2020)Tampuu, Matiisen, Semikin, Fishman and Muhammad}]{tampuu2020survey}
\bibinfo{author}{Tampuu, A.}, \bibinfo{author}{Matiisen, T.}, \bibinfo{author}{Semikin, M.}, \bibinfo{author}{Fishman, D.}, \bibinfo{author}{Muhammad, N.}, \bibinfo{year}{2020}.
\newblock \bibinfo{title}{A survey of end-to-end driving: Architectures and training methods}.
\newblock \bibinfo{journal}{IEEE Transactions on Neural Networks and Learning Systems} \bibinfo{volume}{33}, \bibinfo{pages}{1364--1384}.
\bibitem[{Tang et~al.(2024)Tang, Srishankar, Martin and Tomizuka}]{tang2024grounded}
\bibinfo{author}{Tang, C.}, \bibinfo{author}{Srishankar, N.}, \bibinfo{author}{Martin, S.}, \bibinfo{author}{Tomizuka, M.}, \bibinfo{year}{2024}.
\newblock \bibinfo{title}{Grounded relational inference: Domain knowledge driven explainable autonomous driving}.
\newblock \bibinfo{journal}{IEEE Transactions on Intelligent Transportation Systems} .
\bibitem[{Tian et~al.(2024)Tian, Zhou, Han and Lang}]{tian2024robust}
\bibinfo{author}{Tian, D.}, \bibinfo{author}{Zhou, J.}, \bibinfo{author}{Han, X.}, \bibinfo{author}{Lang, P.}, \bibinfo{year}{2024}.
\newblock \bibinfo{title}{Robust platoon control of mixed autonomous and human-driven vehicles for obstacle collision avoidance: A cooperative sensing-based adaptive model predictive control approach}.
\newblock \bibinfo{journal}{Engineering} .
\bibitem[{Treiber et~al.(2000)Treiber, Hennecke and Helbing}]{treiber2000congested}
\bibinfo{author}{Treiber, M.}, \bibinfo{author}{Hennecke, A.}, \bibinfo{author}{Helbing, D.}, \bibinfo{year}{2000}.
\newblock \bibinfo{title}{Congested traffic states in empirical observations and microscopic simulations}.
\newblock \bibinfo{journal}{Physical Review E} \bibinfo{volume}{62}, \bibinfo{pages}{1805}.
\bibitem[{Wang et~al.(2024)Wang, He, Fan, Li, Chen and Zhang}]{wang2024driving}
\bibinfo{author}{Wang, Y.}, \bibinfo{author}{He, J.}, \bibinfo{author}{Fan, L.}, \bibinfo{author}{Li, H.}, \bibinfo{author}{Chen, Y.}, \bibinfo{author}{Zhang, Z.}, \bibinfo{year}{2024}.
\newblock \bibinfo{title}{Driving into the future: Multiview visual forecasting and planning with world model for autonomous driving}, in: \bibinfo{booktitle}{IEEE/CVF Conference on Computer Vision and Pattern Recognition (CVPR)}, pp. \bibinfo{pages}{14749--14759}.
\bibitem[{Wen et~al.(2023a)Wen, Fu, Li, Cai, Ma, Cai, Dou, Shi, He and Qiao}]{wen2023dilu}
\bibinfo{author}{Wen, L.}, \bibinfo{author}{Fu, D.}, \bibinfo{author}{Li, X.}, \bibinfo{author}{Cai, X.}, \bibinfo{author}{Ma, T.}, \bibinfo{author}{Cai, P.}, \bibinfo{author}{Dou, M.}, \bibinfo{author}{Shi, B.}, \bibinfo{author}{He, L.}, \bibinfo{author}{Qiao, Y.}, \bibinfo{year}{2023}a.
\newblock \bibinfo{title}{{DiLu}: A knowledge-driven approach to autonomous driving with large language models}.
\newblock \bibinfo{journal}{arXiv preprint arXiv:2309.16292} .
\bibitem[{Wen et~al.(2023b)Wen, Fu, Mao, Cai, Dou, Li and Qiao}]{wenl2023limsim}
\bibinfo{author}{Wen, L.}, \bibinfo{author}{Fu, D.}, \bibinfo{author}{Mao, S.}, \bibinfo{author}{Cai, P.}, \bibinfo{author}{Dou, M.}, \bibinfo{author}{Li, Y.}, \bibinfo{author}{Qiao, Y.}, \bibinfo{year}{2023}b.
\newblock \bibinfo{title}{{LimSim}: A long-term interactive multi-scenario traffic simulator}, in: \bibinfo{booktitle}{IEEE International Conference on Intelligent Transportation Systems (ITSC)}, \bibinfo{organization}{IEEE}. pp. \bibinfo{pages}{1255--1262}.
\bibitem[{Xu et~al.(2017)Xu, Gao, Yu and Darrell}]{xu2017end}
\bibinfo{author}{Xu, H.}, \bibinfo{author}{Gao, Y.}, \bibinfo{author}{Yu, F.}, \bibinfo{author}{Darrell, T.}, \bibinfo{year}{2017}.
\newblock \bibinfo{title}{End-to-end learning of driving models from large-scale video datasets}, in: \bibinfo{booktitle}{IEEE Conference on Computer Vision and Pattern Recognition (CVPR)}, pp. \bibinfo{pages}{2174--2182}.
\bibitem[{Xu et~al.(2023)Xu, Xiang, Han, Xia, Meng, Chen, Correa-Jullian and Ma}]{xu2023opencda}
\bibinfo{author}{Xu, R.}, \bibinfo{author}{Xiang, H.}, \bibinfo{author}{Han, X.}, \bibinfo{author}{Xia, X.}, \bibinfo{author}{Meng, Z.}, \bibinfo{author}{Chen, C.J.}, \bibinfo{author}{Correa-Jullian, C.}, \bibinfo{author}{Ma, J.}, \bibinfo{year}{2023}.
\newblock \bibinfo{title}{The opencda open-source ecosystem for cooperative driving automation research}.
\newblock \bibinfo{journal}{IEEE Transactions on Intelligent Vehicles} \bibinfo{volume}{8}, \bibinfo{pages}{2698--2711}.
\bibitem[{Xu et~al.(2024)Xu, Zhang, Xie, Zhao, Guo, Wong, Li and Zhao}]{xu2024drivegpt4}
\bibinfo{author}{Xu, Z.}, \bibinfo{author}{Zhang, Y.}, \bibinfo{author}{Xie, E.}, \bibinfo{author}{Zhao, Z.}, \bibinfo{author}{Guo, Y.}, \bibinfo{author}{Wong, K.Y.K.}, \bibinfo{author}{Li, Z.}, \bibinfo{author}{Zhao, H.}, \bibinfo{year}{2024}.
\newblock \bibinfo{title}{{DriveGPT4}: Interpretable end-to-end autonomous driving via large language model}.
\newblock \bibinfo{journal}{IEEE Robotics and Automation Letters} .
\bibitem[{Yan et~al.(2024)Yan, Pi, Guo, Luo, Dou, Deng, Huang, Fu, Wen, Cai et~al.}]{yan2024oasim}
\bibinfo{author}{Yan, G.}, \bibinfo{author}{Pi, J.}, \bibinfo{author}{Guo, J.}, \bibinfo{author}{Luo, Z.}, \bibinfo{author}{Dou, M.}, \bibinfo{author}{Deng, N.}, \bibinfo{author}{Huang, Q.}, \bibinfo{author}{Fu, D.}, \bibinfo{author}{Wen, L.}, \bibinfo{author}{Cai, P.}, et~al., \bibinfo{year}{2024}.
\newblock \bibinfo{title}{{OASim}: an open and adaptive simulator based on neural rendering for autonomous driving}.
\newblock \bibinfo{journal}{arXiv preprint arXiv:2402.03830} .
\bibitem[{Yang et~al.(2024)Yang, Wen, Ma, Mei, Li, Wei, Lei, Fu, Cai, Dou et~al.}]{yang2024drivearena}
\bibinfo{author}{Yang, X.}, \bibinfo{author}{Wen, L.}, \bibinfo{author}{Ma, Y.}, \bibinfo{author}{Mei, J.}, \bibinfo{author}{Li, X.}, \bibinfo{author}{Wei, T.}, \bibinfo{author}{Lei, W.}, \bibinfo{author}{Fu, D.}, \bibinfo{author}{Cai, P.}, \bibinfo{author}{Dou, M.}, et~al., \bibinfo{year}{2024}.
\newblock \bibinfo{title}{{DriveArena}: A closed-loop generative simulation platform for autonomous driving}.
\newblock \bibinfo{journal}{arXiv preprint arXiv:2408.00415} .
\bibitem[{Yurtsever et~al.(2020)Yurtsever, Lambert, Carballo and Takeda}]{yurtsever2020survey}
\bibinfo{author}{Yurtsever, E.}, \bibinfo{author}{Lambert, J.}, \bibinfo{author}{Carballo, A.}, \bibinfo{author}{Takeda, K.}, \bibinfo{year}{2020}.
\newblock \bibinfo{title}{A survey of autonomous driving: Common practices and emerging technologies}.
\newblock \bibinfo{journal}{IEEE Access} \bibinfo{volume}{8}, \bibinfo{pages}{58443--58469}.
\bibitem[{Zeng et~al.(2019)Zeng, Luo, Suo, Sadat, Yang, Casas and Urtasun}]{zeng2019end}
\bibinfo{author}{Zeng, W.}, \bibinfo{author}{Luo, W.}, \bibinfo{author}{Suo, S.}, \bibinfo{author}{Sadat, A.}, \bibinfo{author}{Yang, B.}, \bibinfo{author}{Casas, S.}, \bibinfo{author}{Urtasun, R.}, \bibinfo{year}{2019}.
\newblock \bibinfo{title}{End-to-end interpretable neural motion planner}, in: \bibinfo{booktitle}{IEEE/CVF Conference on Computer Vision and Pattern Recognition (CVPR)}, pp. \bibinfo{pages}{8660--8669}.
\bibitem[{Zhang et~al.(2022)Zhang, Guo, Zeng, Xiong, Dai, Hu, Ren and Urtasun}]{zhang2022rethinking}
\bibinfo{author}{Zhang, C.}, \bibinfo{author}{Guo, R.}, \bibinfo{author}{Zeng, W.}, \bibinfo{author}{Xiong, Y.}, \bibinfo{author}{Dai, B.}, \bibinfo{author}{Hu, R.}, \bibinfo{author}{Ren, M.}, \bibinfo{author}{Urtasun, R.}, \bibinfo{year}{2022}.
\newblock \bibinfo{title}{Rethinking closed-loop training for autonomous driving}, in: \bibinfo{booktitle}{European Conference on Computer Vision (ECCV)}, \bibinfo{organization}{Springer}. pp. \bibinfo{pages}{264--282}.
\bibitem[{Zhao et~al.(2024)Zhao, Zhao, Deng, Wang, Zhang, Zheng, Cao, Nan, Lian and Burke}]{zhao2024autonomous}
\bibinfo{author}{Zhao, J.}, \bibinfo{author}{Zhao, W.}, \bibinfo{author}{Deng, B.}, \bibinfo{author}{Wang, Z.}, \bibinfo{author}{Zhang, F.}, \bibinfo{author}{Zheng, W.}, \bibinfo{author}{Cao, W.}, \bibinfo{author}{Nan, J.}, \bibinfo{author}{Lian, Y.}, \bibinfo{author}{Burke, A.F.}, \bibinfo{year}{2024}.
\newblock \bibinfo{title}{Autonomous driving system: A comprehensive survey}.
\newblock \bibinfo{journal}{Expert Systems with Applications} \bibinfo{volume}{242}, \bibinfo{pages}{122836}.
\bibitem[{Zheng et~al.(2025)Zheng, Song, Guo, Zhang and Chen}]{zheng2025genad}
\bibinfo{author}{Zheng, W.}, \bibinfo{author}{Song, R.}, \bibinfo{author}{Guo, X.}, \bibinfo{author}{Zhang, C.}, \bibinfo{author}{Chen, L.}, \bibinfo{year}{2025}.
\newblock \bibinfo{title}{{GenAD}: Generative end-to-end autonomous driving}, in: \bibinfo{booktitle}{European Conference on Computer Vision (ECCV)}, \bibinfo{organization}{Springer}. pp. \bibinfo{pages}{87--104}.
\bibitem[{Zhu and Zhao(2021)}]{zhu2021survey}
\bibinfo{author}{Zhu, Z.}, \bibinfo{author}{Zhao, H.}, \bibinfo{year}{2021}.
\newblock \bibinfo{title}{A survey of deep {RL} and {IL} for autonomous driving policy learning}.
\newblock \bibinfo{journal}{IEEE Transactions on Intelligent Transportation Systems} \bibinfo{volume}{23}, \bibinfo{pages}{14043--14065}.

\end{thebibliography}

\end{document}